\documentclass{article}
\newcommand{\tblwidth}{\textwidth}

\usepackage{graphicx}%
\graphicspath{{figs/}{./}}
\usepackage{multirow}%
\usepackage{amsmath,amssymb,amsfonts}%
\usepackage{amsthm}%
\usepackage{mathrsfs}%
\usepackage[title]{appendix}%
\usepackage{xcolor}%
\usepackage{textcomp}%
\usepackage{manyfoot}%
\usepackage{booktabs}%
\usepackage{algorithm}%
\usepackage{algorithmicx}%
\usepackage{algpseudocode}%
\usepackage{listings}%
\usepackage{subcaption}
\usepackage{wrapfig}
\usepackage{tabularx}
\usepackage[numbers]{natbib}
\usepackage{url}
\usepackage{hyperref}
\usepackage{PRIMEarxiv}

\title{Lightweight CNN-Based Anomaly Detection for High Voltage Converter Modulators in the Spallation Neutron Source}

\author{
Alberto D. Cencillo$^{1}$\\
\texttt{albertodiazcen@ugr.es} \\
\And
Leonardo Concepción$^{1}$\\
\texttt{leonar2cp@ugr.es}\\
\And
Julián Luengo$^{2,1}$\\
\texttt{julianlm@decsai.ugr.es}\\[1.0em]
\And
Isaac Triguero$^{2,1}$\\
\texttt{triguero@decsai.ugr.es}\\
$^1$Andalusian Research Institute in Data Science and Computational Intelligence (DaSCI)\\
$^2$Department of Computer Science and Artificial Intelligence (DECSAI),\\
University of Granada, Granada, 18071, Spain
}

\begin{document}
\maketitle


\begin{abstract}
Time series anomaly detection is central to monitoring industrial and scientific equipment, where anomalous sensor readings can precede failures. As such equipment usually carries several sensors, a detector must also decide how to model the dependencies between channels. In many physical systems, normal behaviour is defined by how groups of sensors evolve together, and an anomaly can appear as the coupling or decoupling of sensors while each signal remains normal. Anomalies in popular public benchmarks, however, are mostly visible in single channels, so these benchmarks cannot show whether modelling inter-channel dependencies improves detection. We study this on the High Voltage Converter Modulators (HVCMs) of the Spallation Neutron Source (SNS), whose fourteen sensors are physically coupled and whose faults are labelled by type. Keeping a convolutional neural network (CNN) fixed, we vary only whether its blocks combine channels before or after temporal filtering, and measure detection for each SNS subsystem and fault family. The effect of this choice depends on how visible each fault family is on a single channel. Families clearly visible on one channel are detected equally well by all variants, whereas weakly separable families, such as flux faults, gain consistently when channels are combined first. Modelling the dependencies between channels therefore helps only where individual channels do not already reveal the fault. The best detector reaches an average area under the precision--recall curve of 0.816, competitive with published HVCM detectors at roughly half the parameters of a standard CNN.

\end{abstract}

\keywords{anomaly detection, multivariate time series, channel dependence, particle accelerators, high voltage converter modulator}

\maketitle

\clearpage
\section{Introduction}\label{sec:intro}

Anomaly detection in time series is the task of identifying observations, or segments of a series, that depart from the normal behaviour of the process that generated them \citep{blazquez2021review}. It is a central part of condition monitoring in industrial plants, energy systems, transport and large scientific facilities, where an anomaly in the recorded signals can precede a failure, and early detection allows operators to act before an unscheduled stop \citep{zamanzadeh2024survey}. In most of these settings a process is observed by several sensors at once, so the data form a multivariate time series in which each channel records one physical quantity. Deep learning has become the dominant approach to this problem \citep{zamanzadeh2024survey}, with architectures that range from recurrent and convolutional autoencoders, such as the multi-scale convolutional recurrent variational autoencoder of \citet{xie2023mscrvae}, to transformers and graph neural networks.

Every multivariate detector has to decide how to use the relationships between its channels. Channel-dependent (CD) models process all channels jointly, so that any learned feature can combine information from several sensors. Graph-based detectors are an explicit form of this approach. The graph deviation network (GDN) of \citet{deng2021gdn} learns a graph of the dependencies between sensors and flags deviations from it, and \citet{qiao2026ccgdn} combine graph attention with cross-correlation analysis to capture both synchronous and time-lagged associations between entities. Channel-independent (CI) models, in contrast, process each channel separately with a shared network and combine the channels only when the final anomaly score is formed. CD models can, in principle, detect anomalies that appear only in the way channels evolve together, which CI models cannot, whereas CI models are simpler. Hybrid designs combine both strategies, for example the dual-channel transformer of \citet{huang2026dualchannel}, which pairs a CI temporal encoder with a CD encoder of the relationships between variables.

On public benchmarks, however, the evidence has not favoured CD models. In forecasting, the CI model PatchTST \citep{nie2023a} outperformed the transformer models that preceded it, which mix all variables at each time step, and the linear model DLinear \citep{zeng2023dlinear} was competitive with the same transformers. Two recent studies explain this result. \citet{chen2025closer} show that the widely used forecasting benchmarks contain little useful information between variables, and that models relying more on other variables forecast worse. \citet{benchmarks2026univariate} examine eight widely used multivariate anomaly detection benchmarks and find that no labelled anomaly changes the relationship between channels without also producing a deviation in at least one individual channel. On these data, CD detectors show no measurable advantage over CI detectors.

These findings reveal a limitation of the benchmarks, not a property of multivariate anomalies in general. In industrial process monitoring it has long been established that the measured variables are correlated through the process itself, and that some faults appear only as a break in this correlation structure while every variable stays within its own control limits. Such faults cannot be detected with one control chart per variable, which is why multivariate statistical process control monitors the variables jointly \citep{macgregor1995spc}. \citet{qin2003spm} reviews the detection indices developed for this purpose, among them the squared prediction error, which measures departures from the correlation structure observed under normal operation. The same observation motivates recent work on machine fault diagnosis, where the information from a single sensor is insufficient to characterise system-level faults \citep{chen2026dwdf} and multi-channel signals are fused to identify them \citep{sun2026stff}. The evidence that CD models can exploit such dependencies, however, comes from synthetic data. \citet{chen2025closer} and \citet{benchmarks2026univariate} both obtain a CD advantage only when dependencies are planted in the data, and, to our knowledge, the effect of modelling these dependencies has not been examined on real data in which they arise from the physics of the monitored system.

A second open point concerns how the architecture of a detector determines the way it captures the dependencies between channels. Most of the comparisons above contrast different models, which differ in many respects besides their treatment of channels, so the effect of a particular architectural decision on how these dependencies are modelled, and hence on detection, remains largely unknown. Convolutional detectors, which are used in fault diagnosis on multi-sensor signals, as in the convolutional autoencoder of \citet{sun2026stff}, illustrate the point. Whether a convolutional block mixes channels together with, before or after its temporal filtering determines at which stage the relationships between channels enter the representation. A standard convolution learns both operations within one kernel, whereas depthwise-separable designs, used for example in the CI temporal encoder of \citet{huang2026dualchannel}, filter each channel separately and mix the channels in a separate step.

Particle accelerators offer real data in which such dependencies arise from physics. At the SNS of Oak Ridge National Laboratory, an accelerator-driven neutron source used for materials research \citep{mason2006sns}, fifteen HVCMs supply the high-voltage pulses that drive the klystrons of the linear accelerator (linac). They are consistently the second-largest source of unscheduled downtime at the facility \citep{radaideh2022dsp}, individual failures have caused downtimes of a day or more \citep{pappas2021ml}, and their reliability has been the subject of a sustained improvement programme \citep{anderson2014hvcm}. Each HVCM pulse is recorded by fourteen sensors placed at successive stages of the same converter circuit, so the channels are coupled through the circuit, and a public dataset \citep{radaideh2022data} provides normal and pre-fault pulses from all four SNS subsystems, with the fault type of each pre-fault pulse labelled. Previous studies on this dataset have proposed recurrent and convolutional autoencoders \citep{radaideh2022dsp}, a multi-module conditional variational autoencoder (CVAE) \citep{alanazi2023cvae} and supervised classifiers \citep{zhou2024multisensor}, and \citet{zhou2024multisensor} show that fusing the fourteen channels detects faults better than any single channel. These results confirm that the HVCM channels carry complementary information. They were obtained to detect faults on this particular system, however, and do not address how the dependencies between channels should be modelled, which is the open problem outlined above.

This paper brings the two open points together. We use the HVCM dataset as a real system in which the dependencies between channels arise from physics, and on it we isolate one architectural decision in a CNN, the order in which channels are mixed and time is filtered. We build four detectors that share the same backbone, classifier head and training procedure and differ only in the operation inside each convolutional block. One uses a standard convolution, which mixes channels and filters in time within a single kernel. The other three factorise this kernel into a per-channel temporal filter and a cross-channel mixing step, and apply the two in different orders. Because the detectors are otherwise identical, and the two factorised orders have almost the same number of parameters, differences in detection can be attributed to the way they model the dependencies between channels. If the information that distinguishes a fault lies in the relationships between channels, combining channels before filtering them in time should improve detection, whereas if it is visible in individual channels the order should make little difference. Since fault types differ in how visible they are on single channels, we evaluate the detectors on each of the four SNS subsystems and on each fault family, and we relate the effect of the order to how well each family can be separated from normal pulses on a single channel.

The contributions of this paper are the following:
\begin{itemize}
\item A controlled study of how the order of channel mixing and temporal filtering in a fixed CNN affects anomaly detection on a real, physically coupled sensor system.

\item Evidence that the benefit of cross-channel modelling depends on the fault type. Fault families with a strong single-channel signature show no gain, in agreement with the benchmark studies. Flux faults, and insulated-gate bipolar transistor (IGBT) faults in the one subsystem where they are frequent enough to evaluate, show consistent gains that the benchmarks could not have revealed. In practice, this indicates for which faults a multi-channel detector is worth deploying.

\item A parameter-efficient HVCM fault detector that is competitive with published detectors on the four subsystems, together with channel ablations that identify the sensors on which fault detection depends most.
\end{itemize}

The rest of the paper is organised as follows. Section~\ref{sec:hvcm} describes the HVCM and the public dataset. Section~\ref{sec:related} reviews related work on channel dependence in multivariate time series and on anomaly detection for particle accelerators. Section~\ref{sec:method} introduces the detectors, and Section~\ref{sec:experiments} describes the experimental setup. Section~\ref{sec:analysis} reports the results for each subsystem and fault family, the separability analysis and the channel ablations. Section~\ref{sec:conclusion} concludes.

\section{The High Voltage Converter Modulator}\label{sec:hvcm}

\begin{wrapfigure}{r}{0.55\textwidth}
    \centering
    \includegraphics[width=0.55\textwidth]{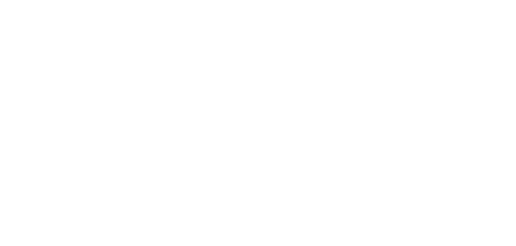}
    \caption{Overview of SNS. Original image by \citet{SNSOverview}.}
    \label{fig:linac}
\end{wrapfigure}

The HVCMs deliver the high-voltage pulses that drive the klystrons of the SNS linac (Figure~\ref{fig:linac}). Each modulator produces these pulses through a chain of stages, in which IGBT switches convert a rectified direct current into an alternating current, pulse transformers raise the voltage, and the output is rectified and filtered before it reaches the klystrons \citep{reass2003hvcm}. The fourteen sensors used in this work measure the currents through the IGBT switches, the magnetic flux in the transformers, and the voltage and current at the modulator output and at the capacitor bank that feeds the chain. Because they observe successive stages of the same circuit, their measurements are not independent, and a fault that develops in one stage can alter the relationships between channels. The fifteen HVCMs are grouped into four subsystems that follow the accelerating structures they power: the radio-frequency quadrupole (RFQ, one modulator), the drift-tube linac (DTL, two), the coupled-cavity linac (CCL, four) and the superconducting linac (SCL, eight). The modulators share the same circuit topology but differ in component values, so their waveforms have a subsystem-specific shape even under normal operation \citep{radaideh2022data}, and we therefore train one detector per subsystem.

We use the public dataset released by \citet{radaideh2022data}, which contains waveforms recorded by the modulator controllers of all fifteen HVCMs between 2020 and 2022. Of the channels recorded by the controllers, the dataset retains the fourteen that domain experts identified as the most informative for fault diagnosis: six IGBT currents (two per phase, labelled A+, A+*, B+, B+*, C+ and C+*), three magnetic fluxes (one per phase, labelled A-Flux, B-Flux and C-Flux), the modulator output voltage and current (Mod-V, Mod-I), the capacitor-bank voltage and current (CB-V, CB-I), and the time derivative of the modulator voltage (DV/DT).

\subsection{The public HVCM dataset}\label{subsec:hvcm-data}

Each record in the dataset is a single pulse of 4500 time steps. For normal records this is the central pulse of the recording, and for faulty records it is the pulse immediately before the fault, so a detector has to recognise a precursor of the fault and not the fault itself. Every pulse is stored as a $4500 \times 14$ array with a binary label (Run or Fault). Faulty pulses also carry a fault-type label that names the component or protection signal involved, such as DV/DT, Flux, IGBT, Driver, SCR (silicon-controlled rectifier) or SNS PPS. These labels follow the categorisation of \citet{alanazi2023cvae}. Table~\ref{tab:dataset} reports the number of pulses per subsystem.

\begin{table}[t]
\caption{HVCM pulse counts per subsystem in the public dataset of
\citet{radaideh2022data}.}\label{tab:dataset}
\centering
\begin{tabular*}{\tblwidth}{@{\extracolsep{\fill}}lrrr@{}}
\toprule
Subsystem & Normal & Faulty & Total \\
\midrule
RFQ   & 690   & 182   & 872 \\
DTL   & 918   & 159   & 1{,}077 \\
CCL   & 1{,}752 & 305 & 2{,}057 \\
SCL   & 3{,}894 & 704 & 4{,}598 \\
\midrule
Total & 7{,}254 & 1{,}350 & 8{,}604 \\
\bottomrule
\end{tabular*}
\end{table}

\begin{figure}
    \centering
    \includegraphics[width=0.99\linewidth]{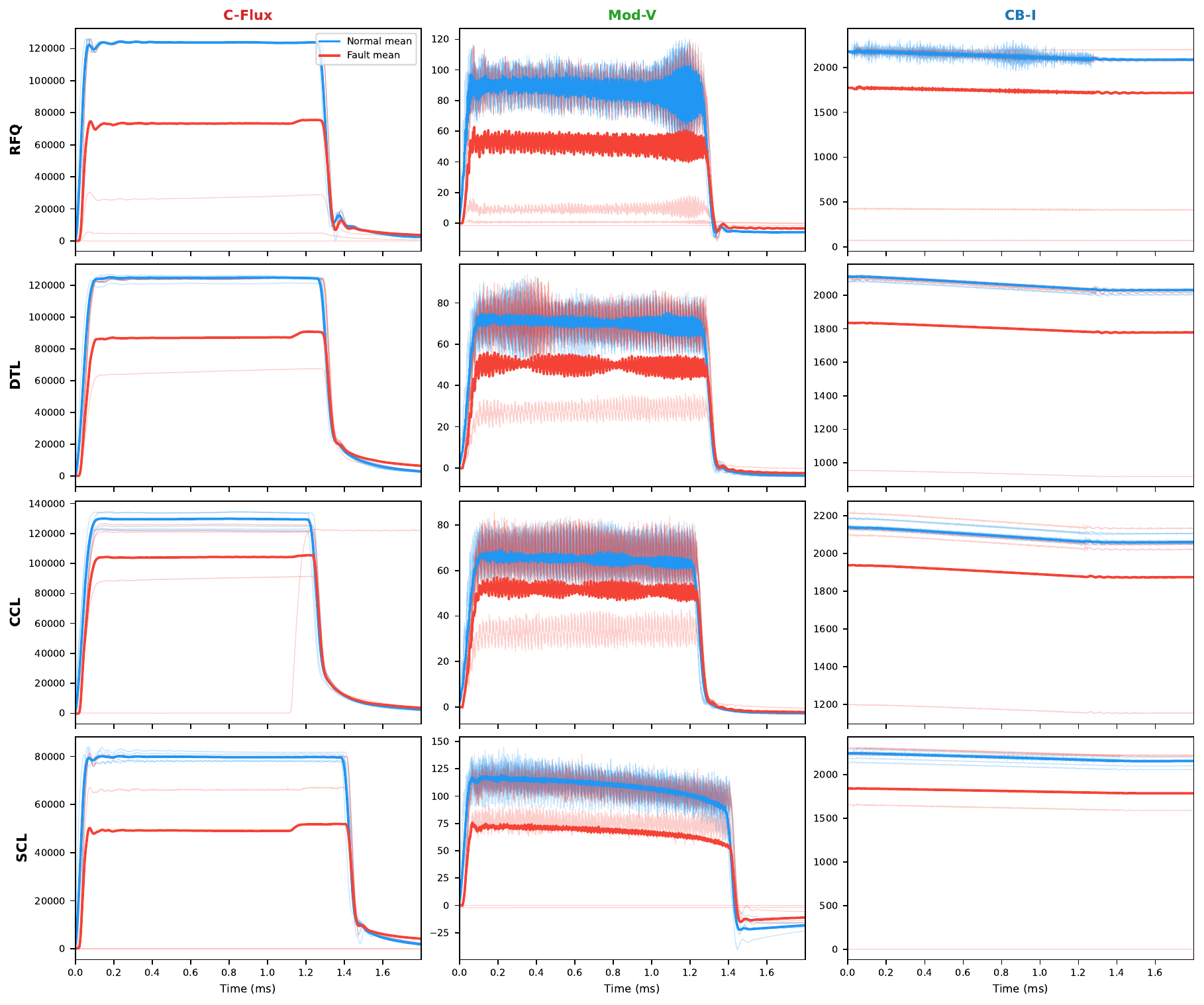}
    \caption{Waveforms of channels C-Flux, Mod-V and CB-I in every one of the four subsystems: RFQ, DTL, CCL and SCL. Pre-fault pulses are shown in red and normal pulses in blue, with the mean waveform of each class overlaid. The degree of class separability differs between subsystems.}
    \label{fig:rfq_ccl}
\end{figure}

Figure~\ref{fig:rfq_ccl} shows normal and pre-fault waveforms of three channels in each subsystem. The figure shows that how well the two classes separate depends on the subsystem. In the RFQ, the C-Flux and Mod-V channels shift clearly between normal and pre-fault pulses, whereas in the CCL the same channels overlap much more. Even within one fault type, pre-fault pulses range from clearly anomalous to almost identical to normal pulses, which \citet{radaideh2022dsp} attribute to different lead times before the fault. As a result, a detector has to cope with a large overlap between the classes, and the evaluation has to use metrics that are robust to class imbalance (Section~\ref{sec:experiments}). The correlations between channels also change between normal and pre-fault pulses, and the change differs between subsystems (Figure~\ref{fig:correlation_study}). The detectors of Section~\ref{sec:method} are designed to test whether modelling these changes improves detection.

\section{Related work}\label{sec:related}
This section introduces the main literature on which this study builds. Section~\ref{subsec:rw-cicd} summarises the main approaches to time series anomaly detection and the work on how multivariate detectors should treat the dependencies between channels, including the evidence obtained on public benchmarks. Section~\ref{subsec:rw-acc} reviews anomaly detection on particle accelerator data, the application domain of this study, and groups the existing detectors by how they treat these dependencies.

\subsection{Time series anomaly detection and channel dependence}\label{subsec:rw-cicd}

Early approaches to time series anomaly detection scored each observation by its deviation from a model of normal behaviour. The model could be an autoregressive integrated moving average (ARIMA) forecaster \citep{yaacob2010arima}, a hidden Markov model \citep{li2017hmm}, a principal component analysis (PCA) subspace \citep{shyu2003pca} or the boundary of a one-class support vector machine (SVM) \citep{ma2003svm}, and general-purpose detectors such as the local outlier factor \citep{breunig2000lof} and the isolation forest \citep{liu2008iforest} have also been widely applied. Deep learning replaced these hand-specified models with learned representations \citep{zamanzadeh2024survey}. Reconstruction-based detectors train an autoencoder on normal data and flag the inputs it reconstructs poorly, with variational \citep{an2015vaeanomaly}, recurrent \citep{provotar2019lstmae} and convolutional recurrent \citep{yin2020iotrnnae} variants, and multi-scale designs that combine these components \citep{xie2023mscrvae}. More recent detectors rely on attention, as in the Anomaly Transformer \citep{xu2022anomalytransformer} and DCdetector \citep{yang2023dcdetector}.

Multivariate detectors also differ in how they use the dependencies between channels. Several CD detectors represent these dependencies explicitly as a graph. MTAD-GAT \citep{zhao2020mtadgat} applies graph attention across variables, GDN learns a graph of sensor relationships and flags deviations from it \citep{deng2021gdn}, \citet{he2025graphgan} combine graph attention with adversarial training, and \citet{qiao2026ccgdn} add cross-correlation analysis to capture time-lagged associations. In forecasting, Crossformer \citep{zhang2023crossformer} and iTransformer \citep{liu2024itransformer} apply attention across variables. At the other end, PatchTST \citep{nie2023a} processes each channel separately with a shared transformer, and DLinear \citep{zeng2023dlinear} applies a linear map along time to each channel. Hybrid designs try to combine the strengths of both. CATCH \citep{catch2025} and the heterogeneity-aware model of \citet{hetero2025kbs} are motivated by the observation that CD models overfit when channels are weakly related, while CI models cannot recover interactions that do exist, and the dual-channel transformer of \citet{huang2026dualchannel} fuses the output of a CI temporal encoder with that of a CD encoder of the relationships between variables.

The detectors above are usually compared on a small set of public benchmarks, whose quality has been questioned. \citet{wu2023current} argue that several popular benchmarks contain anomalies that are trivial to detect or wrongly labelled, and \citet{liu2024tsbad} show that flawed datasets and evaluation measures can make reported progress unreliable. Recent work shows that these benchmarks are also unsuitable for deciding how detectors should treat the channels. In forecasting, PatchTST and DLinear match or outperform the transformer forecasters that preceded them, which mix all variables at each time step, and the explicitly CD transformers have not established a consistent advantage. \citet{chen2025closer} estimated, for each trained forecaster, how much its output depends on the variable being predicted and how much on the others. They found that almost all useful information comes from the variable itself, that models relying more on the other variables forecast worse, and that the average correlation between variables in these datasets is low. In anomaly detection, \citet{benchmarks2026univariate} classified every labelled anomaly in eight common benchmarks according to whether it produces a deviation in a single channel, a change in the relationships between channels, or both, and found no anomaly of the second kind without a deviation of the first. Consequently, a CI detector is sufficient on these benchmarks. Both studies also generated synthetic data with planted dependencies between variables and showed that CD models improve on these data, whereas CI models do not, so the lack of a CD advantage on the public benchmarks is a property of the data and not of the models.

The CD advantage, however, has so far been demonstrated only on synthetic data. In addition, the studies above compare different models, which differ in many respects besides their treatment of channels, so it remains unclear how a specific architectural decision shapes the way a detector captures the dependencies between channels, and in turn its detection performance. Section~\ref{subsec:rw-acc} shows that the same holds for the detectors developed for accelerator data.

\subsection{Anomaly detection for particle accelerators}\label{subsec:rw-acc}

Particle accelerators provide real data from coupled sensors. Each accelerator subsystem is monitored by several sensors whose readings are linked by the physics of the machine, and faults in these subsystems cause unscheduled stops that interrupt the experiments relying on the beam. Neural networks were first applied in this field to accelerator control \citep{nguyen1991slac} and to the modelling of accelerator systems \citep{edelen2016neural}, and have since been used to detect faults and anomalies in the recorded signals. Because these signals come from coupled sensors, detectors for accelerator data face the design decision discussed in Section~\ref{subsec:rw-cicd}, namely how to treat the relationships between channels. Table~\ref{tab:rw} groups the existing detectors according to how they treat the temporal structure within each channel and the relationships between channels, and the following paragraphs discuss the four groups in turn.

The first group in Table~\ref{tab:rw} fixes one of the two axes before learning. Faults in superconducting radio-frequency cavities have been classified with random forests (RF) on engineered features \citep{tennant2020srf}. At the SNS, errant beam pulses have been predicted from beam-current data after spectral feature extraction \citep{rescic2020binary}, an approach later refined by the same authors \citep{rescic2022preemptive}. Faulty beam-position monitors have been identified with a singular value decomposition (SVD) across sensors followed by an isolation forest \citep{fol2020bpm}, and HVCM flux waveforms have been reduced with a discrete cosine transform (DCT) \citep{pappas2021ml}. These pipelines are interpretable and need little data, but they cannot use patterns that lie in the interaction between channels, or temporal features that the chosen transform does not capture.

The second group uses recurrent networks, which learn the temporal features but leave the relationships between channels implicit. \citet{wielgosz2017lhc} monitored the superconducting magnets of the Large Hadron Collider by comparing the output of a recurrent forecaster with the measured signal, and later applied the approach to the magnets of its high-luminosity upgrade \citep{wielgosz2018lhc}. \citet{radaideh2022dsp} used recurrent autoencoders on one HVCM channel at a time. When several signals are available they are usually concatenated into the input vector and their relationships are left to the recurrent state, as in the disruption predictor for fusion plasmas of \citet{kates2019nature}.

The third group mixes all channels from the first layer, either with standard convolutions, as in the cavity fault classifier of \citet{vidyaratne2022srf}, or with fully connected layers, as in the autoencoders of \citet{edelen2021autoencoders}. In both cases the temporal features and the relationships between channels are learned within the same weights. The multi-module CVAE of \citet{alanazi2023cvae} follows this design for the HVCM and adds a conditioning on the modulator to absorb the differences between modules.

The fourth group of Table~\ref{tab:rw} treats the two axes as separate operations. The ConvLSTM autoencoder of \citet{radaideh2022dsp} combines convolution with recurrence, and the supervised model of \citet{zhou2024multisensor} makes the order explicit: a one-dimensional CNN (1D-CNN) first extracts temporal features from each channel separately, and a long short-term memory (LSTM) network then combines the channels. \citet{zhou2024multisensor} found that this fusion of channels detects faults better than any single channel, which is so far the main evidence that the HVCM channels carry complementary information. More recently, \citet{yang2026stcf} detected anomalies in accelerator multi-sensor data with a temporal convolutional encoder and a squeeze-and-excitation (SE) channel-attention block, trained by self-supervised contrastive learning on normal pulses. The order of the encoder and the attention block is fixed, and detection is reported for all faults together.

In summary, previous HVCM work shows that combining channels improves detection \citep{zhou2024multisensor}. Where channels and time are modelled together, they are either learned within the same kernel or combined in one fixed order, and, as the last column of Table~\ref{tab:rw} shows, no study has examined how this choice affects the way the dependencies between channels are captured.

\begin{table*}[t]
\centering
\footnotesize
\caption{Prior work on accelerator data, grouped by how it represents temporal and cross-channel information. The last column records whether the order of cross-channel mixing relative to per-channel temporal filtering is a controlled variable. In every prior approach the two operations are decoupled by a fixed transform, left implicit, learned within a single kernel, or applied in one fixed order. FFT: fast Fourier transform, RNN: recurrent neural network, MLP: multilayer perceptron, AD: anomaly detection.}
\label{tab:rw}
\begin{tabular}{@{}lllll@{}}
\toprule
Representative work & Temporal model & Channel handling & Objective & Mix--filter order \\
\midrule
\multicolumn{5}{@{}l}{\textit{Fixed representation: one axis set before learning}} \\
\citet{tennant2020srf}          & engineered features & single channel                & classification   & - \\
\citet{rescic2022preemptive}    & FFT spectrum + PCA       & single channel        & classification   & - \\
\citet{fol2020bpm}              & per-signal          & SVD (cross-sensor)  & isolation forest & decoupled \\
\citet{pappas2021ml}            & DCT                 & single channel                & thresholding   & - \\
\midrule
\multicolumn{5}{@{}l}{\textit{Adaptive time, implicit channels}} \\
\citet{wielgosz2017lhc}         & LSTM                & single channel      & forecasting      & - \\
\citet{kates2019nature}         & RNN                 & stacked inputs      & classification   & implicit \\
\citet{radaideh2022dsp}         & Bi-LSTM / Bi-GRU    & single channel      & reconstruction   & - \\
\midrule
\multicolumn{5}{@{}l}{\textit{Joint convolution, channel and time fused from the first layer}} \\
\citet{vidyaratne2022srf}       & 1D-CNN              & joint (dense)       & classification   & entangled \\
\citet{edelen2021autoencoders}  & MLP                 & joint (dense)       & reconstruction   & entangled \\
\citet{alanazi2023cvae}         & 1D-CNN              & joint + module cond.& reconstruction   & entangled \\
\midrule
\multicolumn{5}{@{}l}{\textit{Separable but fixed-order}} \\
\citet{radaideh2022dsp}         & ConvLSTM            & single channel & reconstruction   & - \\
\citet{zhou2024multisensor}     & 1D-CNN + LSTM       & per-channel, then fuse & classification & fixed (filter $\to$ mix) \\
\citet{yang2026stcf}            & temporal CNN        & SE channel attention & contrastive AD & fixed \\
\midrule
\textbf{This work}              & 1D-CNN (depthwise-sep.) & dense, factorised   & classification   & \textbf{varied} \\
\bottomrule
\end{tabular}
\end{table*}

\section{Detector design}\label{sec:method}

We treat HVCM fault detection as binary classification. Given a pulse $X \in \mathbb{R}^{C_{in} \times T}$ with $C_{in}=14$ channels and $T$ time steps, the detector predicts whether the next macro-pulse will be normal or faulty. Following \citet{zhou2024multisensor}, we train on normal and faulty pulses with binary labels, a supervised setting that differs from the unsupervised reconstruction approach of the autoencoder baselines (Section~\ref{subsec:exp-baselines}).

To isolate the effect of one architectural decision on how a detector captures the dependencies between channels, all the detectors in this study share the same backbone, classifier head and training procedure, and differ only in the operation inside each convolutional block. This operation is either a standard convolution or one of three factorised forms that separate cross-channel mixing from per-channel temporal filtering and apply the two in different orders. The following subsections describe these operations and relate the two factorised orders to the CI and CD strategies of Section~\ref{subsec:rw-cicd}.

\subsection{Mixing and filtering in a convolutional block}\label{subsec:method-intuition}

\begin{figure}
    \centering
    \begin{subfigure}{0.29\textwidth}
        \centering
        \includegraphics[width=\linewidth]{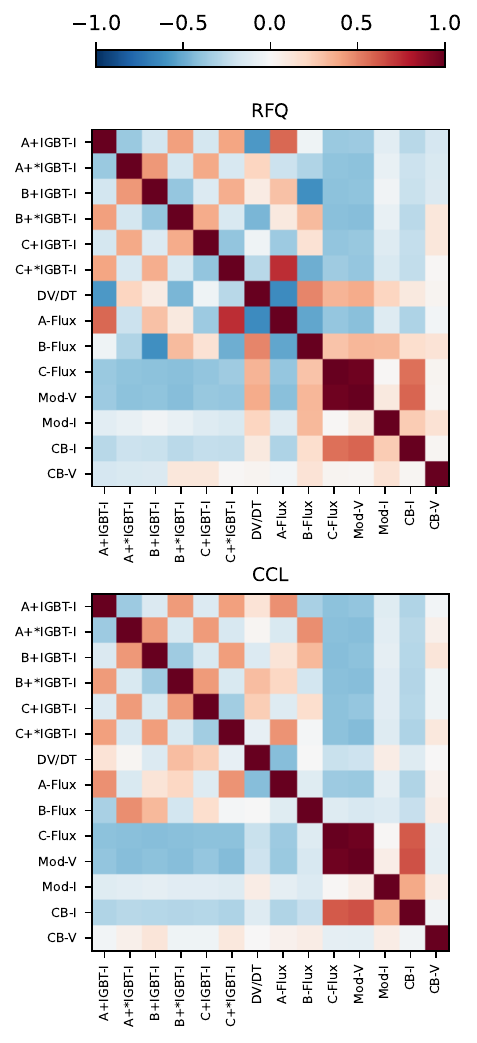}
        \caption{Per-subsystem correlation matrices under normal operation. Some channels, such as C-Flux and Mod-V, are strongly correlated in every subsystem, which reflects the physical coupling between them.}
        \label{fig:corr_normal}
    \end{subfigure}
    \hfill
    \begin{subfigure}{0.29\textwidth}
        \centering
        \includegraphics[width=\linewidth]{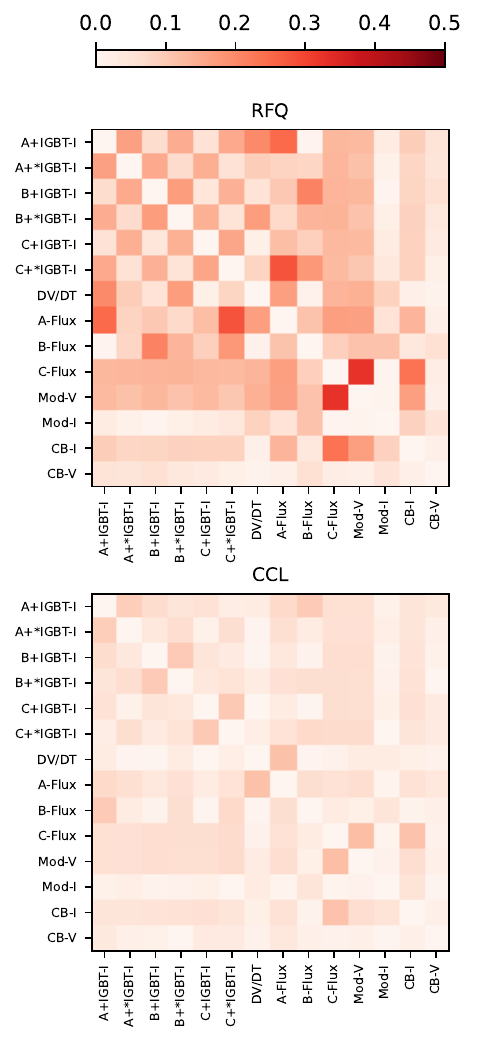}
        \caption{Per-subsystem correlation difference matrices
        ($|\rho_\text{fault} - \rho_\text{normal}|$ for each channel pair).
        RFQ exhibits larger shifts in inter-channel correlation structure than
        CCL.}
        \label{fig:corr_diff}
    \end{subfigure}
    \hfill
    \begin{subfigure}{0.32\textwidth}
        \centering
        \includegraphics[width=\linewidth]{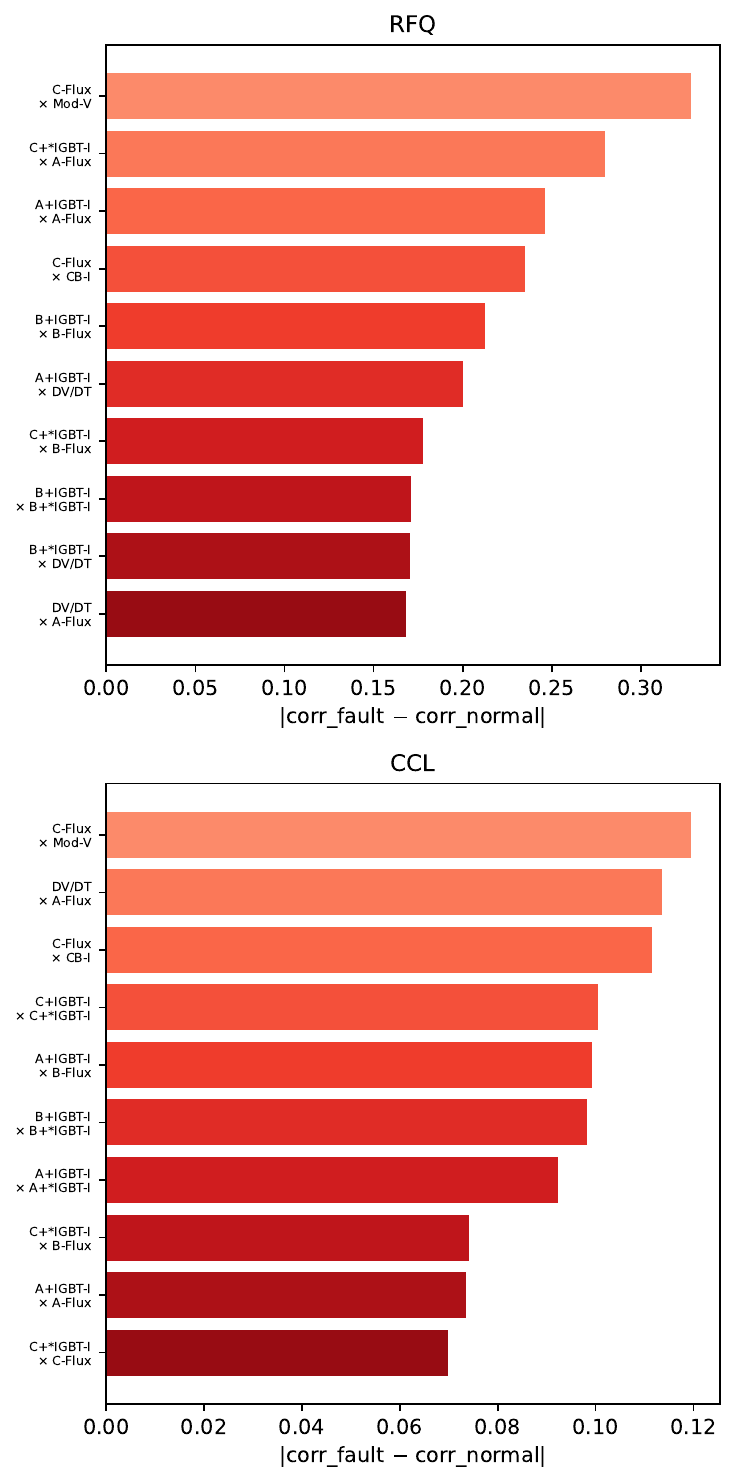}
        \caption{Top channel pairs ranked by correlation difference magnitude.
        RFQ concentrates the largest shifts in a small number of pairs, while
        CCL displays more evenly distributed values.}
        \label{fig:top_pairs}
    \end{subfigure}
    \caption{Cross-channel correlation analysis across RFQ and CCL subsystems.
    Each cell shows the absolute difference in Pearson correlation between the
    normal and pre-fault pulse populations.}
    \label{fig:correlation_study}
\end{figure}

The 14 HVCM channels combine currents, voltages, magnetic fluxes and a derivative signal, with different units, dynamic ranges and noise levels \citep{radaideh2022data}. Fault precursors do not appear in the same way on all channels. Some fault families disturb most channels, others leave a weaker signature on a few, and within one family the strength of the signature varies from pulse to pulse (Figure~\ref{fig:rfq_ccl}). The correlations between channels also change before a fault, and they change differently across subsystems (Figure~\ref{fig:correlation_study}). In the RFQ and SCL the largest changes are concentrated in a few channel pairs, while in the CCL and DTL smaller changes are spread more evenly. Hence, a detector for these data may need both the temporal shape of each channel and the relationships between channels. A standard convolution learns the two within a single kernel from the first layer, so they cannot be varied independently. Factorising the kernel separates them, and the order in which the two factors are applied can then be changed while the rest of the detector is kept fixed.

A standard convolutional block maps $X \mapsto Y \in \mathbb{R}^{C_\text{out}\times T}$ through

\begin{equation}\label{eq:conv}
Y_o(t) = \sum_{c=1}^{C_\text{in}} \sum_{j=0}^{k-1} W_{o,c,j}\, X_c(t-j),
\end{equation}

where $W_{o,c,\cdot}$ is the temporal filter through which input channel $c$ feeds output channel $o$. A full convolution leaves $W$ unconstrained, binding channel index and time offset inside one kernel at a cost of
$C_\text{in} C_\text{out} k$ parameters per block. Depthwise-separable convolutions, introduced in Xception \citep{chollet2017xception} and MobileNets \citep{howard2017mobilenets} and adapted to time series in ModernTCN \citep{donghao2024moderntcn}, split this kernel into two operators that act on different axes. The depthwise operator applies one temporal filter per channel and performs no mixing,

\begin{equation}\label{eq:dw}
(\mathcal{D}X)_c(t) = \sum_{j} d_{c,j}\, X_c(t-j),
\qquad d \in \mathbb{R}^{C_{in} \times k},
\end{equation}

and the pointwise ($1{\times}1$) operator recombines channels at each instant with no temporal extent,

\begin{equation}\label{eq:pw}
(\mathcal{P}X)_o(t) = \sum_{c} P_{o,c}\, X_c(t),
\qquad P \in \mathbb{R}^{C_\text{out} \times C_\text{in}}.
\end{equation}

Because $\mathcal{D}$ and $\mathcal{P}$ act on different axes, they can be composed in either order, and this order is the architectural decision studied here. We compare the two orders, and a gated version of one of them, with a \emph{Standard} baseline that uses Equation~\ref{eq:conv}.

\subsection{The DS variant: depthwise-separable extraction}\label{subsec:method-ds}

Depthwise-Separable (\emph{DS}) filters each channel before mixing,

$\mathcal{B}_\text{DS} = \mathcal{P} \circ \mathcal{D}$:
\begin{equation}\label{eq:ds}
(\mathcal{B}_\text{DS}X)_o(t) = \sum_{c} P_{o,c}\,(d_c * X_c)(t),
\qquad W_{o,c,j} = P_{o,c}\, d_{c,j}.
\end{equation}

\begin{figure}
    \centering
    \includegraphics[width=0.95\linewidth]{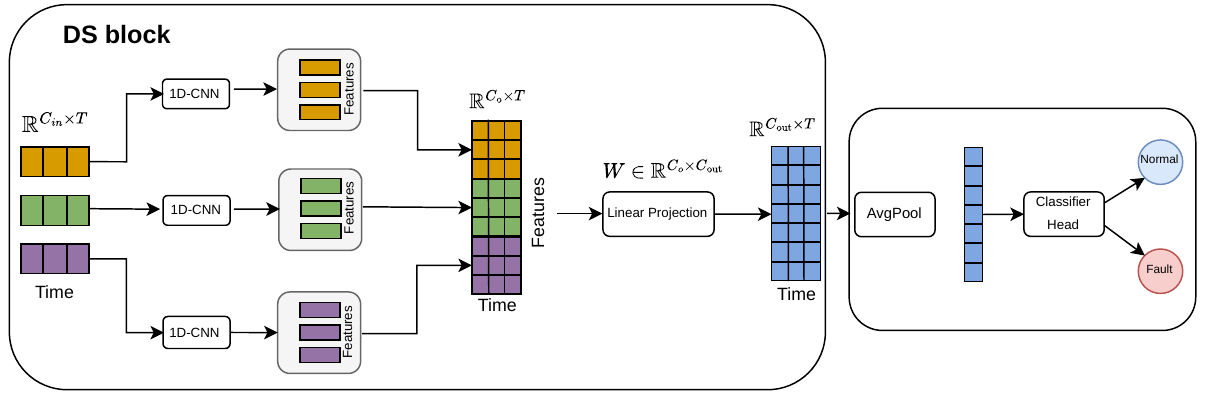}
    \caption{DS architecture.}
    \label{fig:ds}
\end{figure}

The temporal filter $d_c$ is indexed by the \emph{input} channel: every physical sensor keeps its own filter, and mixing acts on already-filtered sensors. A diagram is shown in Figure~\ref{fig:ds}.

DS thus assumes that the most informative low-level features are the shapes of individual channels, for example a flux signal that falls below its normal envelope or an IGBT current that fails to commutate, and it leaves the relationships between channels to the pointwise mixing that follows.

\subsection{The PW-First variant: pointwise-then-depthwise}\label{subsec:method-pw}

Pointwise-First (\emph{PW-First}) reverses the order,

$\mathcal{B}_\text{PW} = \mathcal{D}' \circ \mathcal{P}$:
\begin{equation}\label{eq:pwf}
(\mathcal{B}_\text{PW}X)_o(t) = \sum_{c} P_{o,c}\,(d'_o * X_c)(t),
\qquad W_{o,c,j} = P_{o,c}\, d'_{o,j}.
\end{equation}

\begin{figure}
    \centering
    \includegraphics[width=0.95\linewidth]{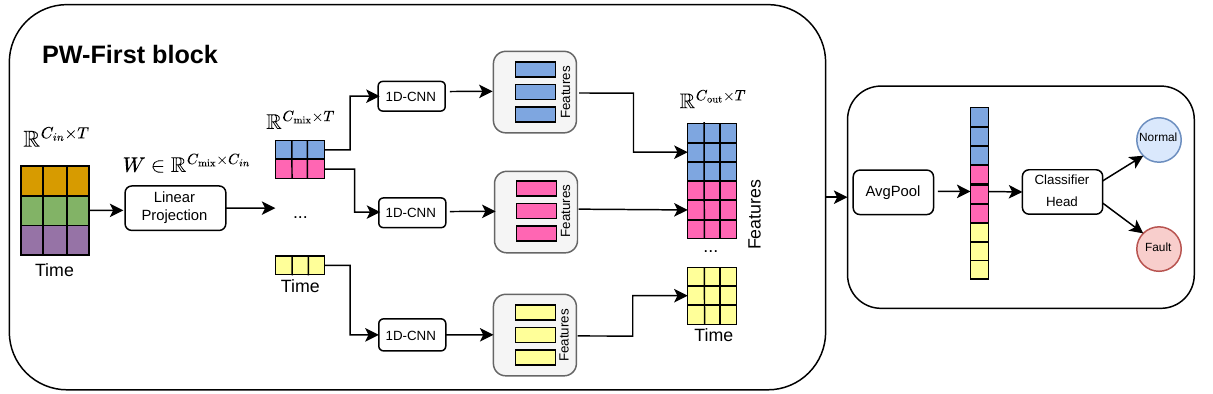}
    \caption{PW-First architecture.}
    \label{fig:pw-first}
\end{figure}

The block first projects the input channels onto the block width with a pointwise $1{\times}1$ convolution, which can be read as a learned linear combination of the raw signals, and then filters each combined channel with a depthwise temporal kernel. The architecture is shown in Figure~\ref{fig:pw-first}.

Equations~\eqref{eq:ds} and~\eqref{eq:pwf} show the structural difference between the two orders. Along the path $c \to o$, the temporal filter is indexed by the input channel $c$ in DS and by the derived channel $o$ in PW-First. In DS every sensor is filtered by its own kernel before mixing. In PW-First the mixing first forms learned combinations of channels, and each combination is then filtered by a single kernel, whatever sensors it contains.

The two orders place the detector at different points between the CI and CD strategies of Section~\ref{subsec:rw-cicd}. In the first block of DS, each physical sensor is filtered in time on its own before any mixing, so the lowest-level temporal features are extracted separately for each channel, which is close to a CI model, and the relationships between channels enter only afterwards. PW-First mixes the channels before any temporal filtering, so every temporal feature is computed from a combination of sensors, as in a CD model. The standard convolution learns the temporal features and the channel combinations jointly. If the information that distinguishes a fault lies in the temporal shape of individual channels, the order should matter little. If it lies in the relationships between channels, for example in the relative phase of two IGBT currents or in the difference between the modulator and capacitor-bank voltages, PW-First can form the relevant combinations before the temporal filters are applied. The correlation changes in Figure~\ref{fig:correlation_study} suggest that HVCM fault precursors may alter such relationships, and Section~\ref{sec:analysis} tests whether the order matters for each fault family.

\subsection{The PW-First+SE variant: PW-First with channel attention}\label{subsec:method-pwse}
The SE block \citep{hu2018squeezeexcitation} is a standard way to recalibrate channels in convolutional networks, and it has been used in fault detection models to capture dependencies between feature channels and suppress uninformative ones \citep{yu2024eeai}. In the \emph{PW-First+SE} variant, we insert it between the two stages of the PW-First block, so that each block becomes pointwise convolution $\rightarrow$ SE $\rightarrow$ depthwise convolution. Let $U = \mathcal{P}X$ be the mixed signal. The gate computes a weight for each channel from a global temporal summary and rescales each derived channel before the depthwise filtering:

\begin{equation}\label{eq:se}
z_o = \frac{1}{T}\sum_t U_o(t), \qquad
s = \sigma\!\bigl(W_2\, \rho(W_1 z)\bigr) \in (0,1)^{C_\text{out}}, \qquad
\tilde{U}_o(t) = s_o\, U_o(t),
\end{equation}

with reduction ratio $r=4$ (so $W_1 \in \mathbb{R}^{(C_\text{out}/r) \times C_\text{out}}$), $\rho$ a rectified linear unit (ReLU) and $\sigma$ a sigmoid. The effective filter becomes $W_{o,c,j}(X) = s_o(X)\, P_{o,c}\, d'_{o,j}$. The only change from Equation~\eqref{eq:pwf} is that $W$ now depends on $X$. Standard, DS and PW-First fix their mixing during training, whereas PW-First+SE keeps the order of PW-First and rescales the mixed channels of each pulse according to the pulse itself. The gate creates no new combinations. It only reweights existing ones, and it adds $2C_{\mathrm{out}}^2/r$ parameters per block. In our backbone, $s$ has a different length in each block and indexes derived channels, not the 14 physical inputs.

\subsection{Parameter cost}

Per block, Standard has $C_{\mathrm{in}}C_{\mathrm{out}}k$ parameters, while the factorised variants have $C_{\mathrm{in}}C_{\mathrm{out}}+Ck$, roughly $k$ times fewer for wide blocks. The SE gate adds $2C_{\mathrm{out}}^2/r$ parameters per block, so all three factorised variants remain smaller than the Standard baseline.

\begin{table}[t]
\caption{Shared training hyperparameters for the Standard, DS, PW-First and PW-First+SE variants.}\label{tab:hparams}
\centering
\begin{tabular}{@{}ll@{}}
\toprule
Hyperparameter & Value \\
\midrule
Input shape              & $14 \times T$ (channels $\times$ time steps) \\
Number of convolutional blocks & $4$ \\
Output channels / block  & $(32, 64, 128, 64)$ \\
Depthwise kernel size    & $(7, 5, 3, 3)$ \\
Pooling                  & Max pooling (size 2) after blocks 1--3; adaptive average pooling after block~4 \\
Activation               & ReLU after batch normalisation \\
Classifier head          & Dropout (rate 0.3) and a linear layer with a binary output \\
Loss                     & Weighted binary cross-entropy, $w^{+} = |\text{Normal}|/|\text{Fault}|$ per subsystem \\
Optimiser                & Adam, learning rate $10^{-3}$, weight decay $10^{-4}$, cosine annealing \\
Batch size               & $64$ \\
Epochs / early stopping  & Up to 100; patience 20 on validation loss \\
Validation split         & $15$\% of training set (early stopping only) \\
Normalisation            & Per-channel z-score from training-set statistics \\
Random seed              & 15 seeds; mean reported \\
\bottomrule
\end{tabular}
\end{table}

\section{Experimental setup}\label{sec:experiments}

We first compare the four detectors of Section~\ref{sec:method} on each subsystem under a common protocol, so that differences between them can be attributed to the order of channel mixing and temporal filtering. We then break this comparison down by fault family and relate it to how visible each family is on a single channel. Finally, we compare the detectors with published HVCM detectors, whose results are taken from the original papers. 

\subsection{Implementation details}
All four variants share the backbone summarised in Table~\ref{tab:hparams}, with four convolutional blocks, each followed by batch normalisation, a ReLU and pooling, and a linear classifier on the pooled output. In DS, PW-First and PW-First+SE, the standard convolution of each block is replaced by the corresponding factorised operator of Equations~\eqref{eq:ds}--\eqref{eq:se}.

Training minimises a binary cross-entropy loss in which the positive class is weighted by $|\text{Normal}|/|\text{Fault}|$, computed on the training split of each subsystem, so that both classes contribute equally to the loss. Each model is trained with early stopping on the loss of an internal validation set, and the weights with the lowest validation loss are restored before testing. For the threshold-dependent metrics, the decision threshold of each model is the point of its test precision--recall curve that maximises F1 (Section~\ref{subsec:exp-metrics}). Table~\ref{tab:hparams} lists the remaining hyperparameters.

\subsection{Data partition and protocol}
Each of the four subsystems is treated as a separate benchmark with its own stratified 70/30 train/test split at the pulse level, following \citet{zhou2024multisensor}. One model is trained per variant, subsystem and seed, with 15 seeds, which gives 60 models per variant. Each model is evaluated only on the test split of its own subsystem, and no model is trained on more than one subsystem. Subsystem-level results are the mean over the 15 seeds, and pooled results are the mean over all 60 models. Each seed fixes both the data split and the initialisation, so the 15 runs of a subsystem are re-splits of the same pulses, which the statistical protocol below takes into account. Within each training split, 15\% of the pulses are held out as a validation set used only for early stopping. Each channel is standardised with the mean and standard deviation of the training pulses of its subsystem, and the same statistics are applied to the validation and test pulses. Channels with a standard deviation below $10^{-8}$ are given unit variance.

\subsection{Evaluation metrics}\label{subsec:exp-metrics}

Faults make up only 15--21\% of the pulses in each subsystem (Table~\ref{tab:dataset}), so accuracy is misleading: a constant predictor that always outputs Run reaches 79--85\% accuracy without detecting any fault. We therefore report seven metrics:

\begin{itemize}
\item \textbf{AUC-PR}, the area under the precision--recall curve, is the primary metric at the subsystem level. It is robust to class imbalance and requires no threshold.
\item \textbf{AUC-ROC}, the area under the receiver operating characteristic curve, is reported for comparison with the autoencoder baselines (Section~\ref{subsec:exp-baselines}). It is a secondary metric at the subsystem level, because it can overstate performance on imbalanced anomaly detection data \citep{liu2024tsbad}, and it is the per-family metric for the reason given in Section~\ref{subsec:exp-family}.
\item \textbf{Accuracy} is reported only for comparison with \citet{zhou2024multisensor}.
\item \textbf{Precision} $= TP / (TP + FP)$, where $TP$ and $FP$ are the numbers of true and false positives, evaluated at the F1-maximising threshold.
\item \textbf{Recall} $= TP / (TP + FN)$, where $FN$ is the number of false negatives, evaluated at the same threshold.
\item \textbf{F1} is the harmonic mean of precision and recall.
\item \textbf{G-Mean} is the geometric mean of recall and specificity, which is sensitive to performance on the minority class.
\end{itemize}

The four threshold-dependent metrics are computed at the threshold that maximises F1 on the precision--recall curve of each model's test predictions. These metrics thus describe the best operating point of each model and are reported for comparison with previous work. AUC-PR and AUC-ROC require no threshold, and all the conclusions of this paper are based on them.

\subsection{Per-fault-family evaluation and single-channel separability}\label{subsec:exp-family}

The dataset labels each faulty pulse with a fault name (Section~\ref{subsec:hvcm-data}). We group the names into families according to the component they refer to: DV/DT, Flux, Driver, SCR and SNS PPS. All remaining names form a Misc group. Misc is not a physical family. Of its 677 pulses, 419 are labelled Fiber Fault and 66 are IGBT faults, and none of the other names has more than 33 pulses. It is reported for completeness, and none of our conclusions depends on it. IGBT faults are analysed separately, for the reasons given below.

Per-family detection is evaluated on the same trained models and test splits as the subsystem-level results. For a family $g$ and one model, the positives are the test pulses of family $g$, the negatives are all normal test pulses of the same subsystem, and we compute the AUC-ROC of the model's scores on this subset. A family with fewer than three positives in a test split is not evaluated on that split. We use AUC-ROC at this level, instead of the AUC-PR used for the subsystem-level results, because the number of positives per split is small (3 to 15 for most family--subsystem pairs) and AUC-PR depends strongly on the proportion of positives, which would make its values incomparable across families. The cross-channel effect for a family is the paired difference in per-family AUC-ROC between PW-First and DS on the same split and seed, since these two detectors differ only in the order of mixing and filtering. Differences involving PW-First+SE are also reported.

IGBT faults are analysed separately because their 66 pulses are unevenly distributed: none in the RFQ, 4 in the DTL, 40 in the CCL and 22 in the SCL. A per-family estimate is possible only on the CCL (6 to 17 positives per test split) and the SCL (1 to 10), and these two subsystems are reported separately. In the main evaluation IGBT faults belong to the Misc group, so their per-family results come from a second training run with the same protocol and data splits, whose subsystem-level AUC-PR agrees with the main run to within 0.02.

To explain the per-family results, we measure how visible each family is on a single channel. For every family and each of the 14 channels, we compute a Welch $t$-statistic that compares the per-pulse mean amplitude of the family's pulses with that of all normal pulses, pooling the pulses of the four subsystems. Because the $t$-statistic grows with the number of pulses in a family, we also report a measure that does not depend on sample size, the best AUC-ROC that can be obtained by ranking the pulses on the per-pulse mean of a single channel, denoted $\mathrm{AUC}_1$. Both measures are given in Table~\ref{tab:welch}.

\subsection{Statistical protocol}\label{subsec:exp-stats}

The design of the experiments determines how differences can be tested. Each seed fixes both the train/test split and the initialisation, so the 15 runs on one subsystem are re-splits of the same pulses. They are paired across variants but are not independent replicates, and pooling the 60 runs of the four subsystems into a single test would treat re-splits of four datasets as 60 independent observations. In addition, with only four subsystems, a claim about the detectors across subsystems rests on four units and cannot be tested meaningfully.

We therefore test differences within each subsystem and only describe them across subsystems. Within a subsystem, adjacent variants (Standard and DS, DS and PW-First, PW-First and PW-First+SE) and the end-to-end contrast (Standard and PW-First+SE) are compared with two-sided paired Wilcoxon signed-rank tests over the 15 seeds, with a Holm correction across the three adjacent comparisons. Per-family differences are tested in the same way within each subsystem, and their means over subsystems are reported as descriptive summaries. We use non-parametric tests because the paired differences are not necessarily normally distributed and the number of seeds is small. Across subsystems and families, we report the direction, size and consistency of the effects without formal tests.

\subsection{Baselines}\label{subsec:exp-baselines}

The comparisons that support the conclusions of this paper are those between the four detectors. Published HVCM detectors are included for context, and they fall into two groups according to how closely their protocol matches ours.

The first group uses the same protocol. \citet{zhou2024multisensor} train supervised classifiers on all 14 channels of each subsystem, with a 70/30 split and both normal and faulty pulses, as we do. They report classical models, namely $k$-nearest neighbours (KNN), RF and SVM, a 1D-CNN, an LSTM, and a CNN+LSTM with multi-sensor fusion, which performs best. The second group is unsupervised and trained on normal pulses only. The recurrent autoencoders of \citet{radaideh2022dsp}, based on bidirectional LSTM, bidirectional gated recurrent unit (GRU) and ConvLSTM layers, use only the C-Flux channel of the RFQ subsystem. The CVAE of \citet{alanazi2023cvae}, with one-hot module conditioning, three 1D-CNN blocks in the encoder and decoder and a 512-dimensional latent space, uses all 14 channels and all modulators, and reports AUC-ROC per fault family based on the Mod-V reconstruction error. Since these detectors do not see fault labels during training, their results provide context and are not directly comparable with ours.

All baseline results are taken from the original papers and were not re-run under our protocol, so this comparison is indicative.

\section{Results and analysis}\label{sec:analysis}

The results are presented in the order of Section~\ref{sec:experiments}. Section~\ref{subsec:analysis-overall} examines whether the order of channel mixing and temporal filtering changes detection in each subsystem, and Section~\ref{subsec:analysis-significance} tests these differences. Section~\ref{subsec:analysis-family} breaks the results down by fault family and relates them to single-channel separability. Section~\ref{subsec:analysis-corr} examines which channels the trained detectors rely on, and Section~\ref{subsec:analysis-detection-hvcm} compares the detectors with published HVCM detectors. The four variants form a sequence in which each step changes one factor: from Standard to DS the convolution is factorised, from DS to PW-First the order of mixing and filtering is reversed, which moves the lowest-level processing from close to a CI model to a CD one (Section~\ref{subsec:method-pw}), and from PW-First to PW-First+SE the SE gate is added.

\subsection{Overall detection performance}\label{subsec:analysis-overall}

\begin{table}[t]
\centering
\caption{Trainable parameter count per variant for the shared backbone of Section~\ref{sec:experiments} (four blocks of widths 32, 64, 128, 64 with depthwise kernels 7, 5, 3, 3). The three factorised variants are strictly smaller than the Standard joint-mixing baseline.}
\label{tab:params}
\begin{tabular}{lr}
\toprule
Variant & Parameters \\
\midrule
Standard & 63,169 \\
DS & 20,355 \\
PW-First & 20,641 \\
PW-First+SE & 33,801 \\
\bottomrule
\end{tabular}
\end{table}

\begin{figure}
\centering
\includegraphics[width=0.6\columnwidth]{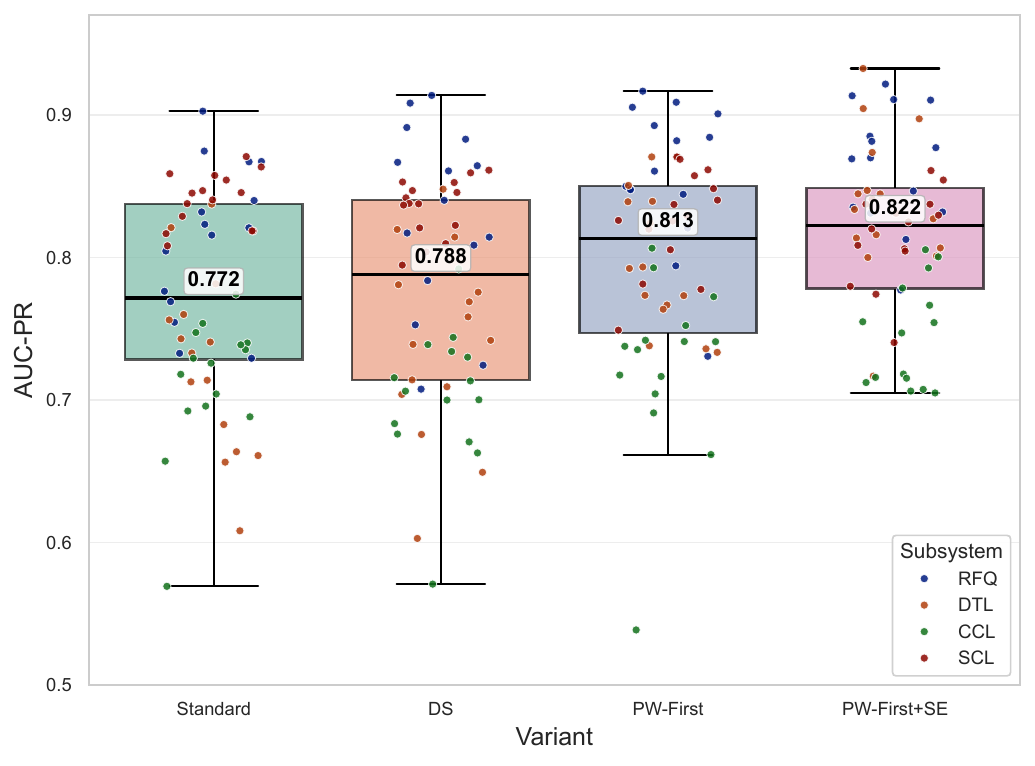}
\caption{Distribution of test AUC-PR across the four subsystems and 15 seeds ($n=60$ per variant). The median ranking Standard $<$ DS $<$ PW-First $<$ PW-First+SE holds, and PW-First+SE additionally has the highest lower quartile and the tightest spread. The boxes show the median and the inter-quartile range (IQR), and the whiskers extend to the most extreme point within $1.5\times$ IQR.}
\label{fig:auprc-boxplot}
\end{figure}

\begin{table}[t]
\caption{Detection performance averaged across the four HVCM subsystems (RFQ, DTL, CCL, SCL). Best value per column in bold.}\label{tab:results-pooled}
\centering
\begin{tabular*}{\tblwidth}{@{\extracolsep{\fill}}lccccccc@{}}
\toprule
Variant & AUC-PR & AUC-ROC & Acc & Prec & Rec & F1 & G-Mean \\
\midrule
Standard    & 0.773 & 0.890 & 0.919 & \textbf{0.844} & 0.629 & 0.714 & 0.781 \\
DS          & 0.777 & 0.898 & 0.918 & 0.835 & 0.637 & 0.714 & 0.784 \\
PW-First    & 0.802 & 0.920 & 0.915 & 0.787 & 0.693 & 0.725 & 0.812 \\
PW-First+SE & \textbf{0.816} & \textbf{0.934} & \textbf{0.919} & 0.792 & \textbf{0.703} & \textbf{0.738} & \textbf{0.820} \\
\bottomrule
\end{tabular*}
\end{table}

\begin{table}[t]
\caption{Per-subsystem AUC-PR. Each cell is the mean over 15 training runs of one model trained on that subsystem only. Best per column in bold.}\label{tab:results-per-subsystem-auprc}
\centering
\begin{tabular*}{\tblwidth}{@{\extracolsep{\fill}}lcccc@{}}
\toprule
Variant & RFQ & DTL & CCL & SCL \\
\midrule
Standard    & 0.814 & 0.725 & 0.711 & \textbf{0.841} \\
DS          & 0.829 & 0.740 & 0.703 & 0.835 \\
PW-First    & 0.859 & 0.798 & 0.723 & 0.827 \\
PW-First+SE & \textbf{0.865} & \textbf{0.837} & \textbf{0.745} & 0.817 \\
\bottomrule
\end{tabular*}
\end{table}

We first compare the number of parameters, so that differences in detection can be separated from differences in capacity. Table~\ref{tab:params} reports the parameter counts for the backbone of Section~\ref{sec:experiments}. The factorised variants are much smaller than the Standard baseline. DS and PW-First have 3.1 times fewer parameters (20,355 and 20,641 against 63,169), and PW-First+SE has 1.9 times fewer despite the added gate. Since DS and PW-First differ by only 1.4\% in parameter count, a difference in performance between them can be attributed to the order in which mixing and temporal filtering are applied.

Table~\ref{tab:results-pooled} reports the metrics averaged over the four subsystems and 15 seeds. On every metric except precision and accuracy, the variants are ranked Standard $<$ DS $<$ PW-First $<$ PW-First+SE. These are averages over subsystems, and Section~\ref{subsec:analysis-significance} tests the differences within each subsystem. The largest gains are on the metrics that depend most on the minority Fault class. From Standard to PW-First+SE, AUC-PR increases by 5.6\% in relative terms, recall by 11.8\% and G-Mean by 5.0\%. Accuracy, which is dominated by the majority Run class, barely changes across variants, which confirms that it is a poor measure of fault detection on this dataset, as \citet{zhou2024multisensor} also noted. The best variant is also one of the smallest, so its improvement cannot be explained by additional capacity. Section~\ref{subsec:analysis-significance} shows which architectural step accounts for it on each subsystem.

Table~\ref{tab:results-per-subsystem-auprc} reports the AUC-PR for each subsystem, and Figure~\ref{fig:auprc-boxplot} shows its distribution over subsystems and seeds. PW-First+SE is the best variant on the RFQ, DTL and CCL, whereas on the SCL the ranking is reversed and Standard performs best. The gain from PW-First to PW-First+SE is largest on the DTL, at about 0.04 AUC-PR. The F1 results in Table~\ref{tab:results-per-subsystem-f1} follow the same pattern.

\subsection{Statistical tests}

\label{subsec:analysis-significance}

The rankings above are based on mean AUC-PR. Following Section~\ref{subsec:exp-stats}, we test the differences within each subsystem, pairing the runs by seed. Table~\ref{tab:steps-persubsystem} reports the three adjacent steps and the end-to-end contrast. The step from Standard to DS is not significant on any subsystem, so the depthwise factorisation alone reduces the number of parameters without improving detection. The step from DS to PW-First is the largest adjacent step on the RFQ, DTL and CCL, and it remains significant after correction on the RFQ and DTL (on the CCL, $p=0.041$ before correction and $0.12$ after). Because DS and PW-First differ only in the order of mixing and filtering, and by 1.4\% in parameters, this step isolates the effect of the order. It replaces low-level processing that is close to a CI model with processing that is CD from the first operation, and it improves detection on the RFQ, DTL and CCL, significantly on the first two. The SE gate adds a further significant improvement on the DTL only. On the SCL every step is zero or negative, and the end-to-end contrast is significantly negative. Across subsystems we only describe the pattern. The effect of the order is present on three subsystems and absent on the fourth, and with four subsystems this count is not tested.
\begin{table}[t]
\caption{Per-subsystem differences in AUC-PR between successive variants, as mean paired differences over the 15 seeds. The three adjacent steps are tested with two-sided paired Wilcoxon signed-rank tests with Holm correction within each subsystem, the end-to-end contrast with an uncorrected two-sided test. Bold: $p<0.01$, underlined: $p<0.05$.}\label{tab:steps-persubsystem}%
\setlength{\tabcolsep}{4pt}
\footnotesize
\begin{tabular*}{\columnwidth}{@{\extracolsep{\fill}}lcccc@{}}
\toprule
Subsystem & Standard $\rightarrow$ DS & DS $\rightarrow$ PW-First & PW-First $\rightarrow$ PW-First+SE & Standard $\rightarrow$ PW-First+SE \\
\midrule
RFQ & $+0.015$ & $\underline{+0.030}$  & $+0.006$     & $\mathbf{+0.051}$ \\
DTL & $+0.015$ & $\mathbf{+0.057}$ & $\underline{+0.040}$ & $\mathbf{+0.112}$ \\
CCL & $-0.009$ & $+0.021$      & $+0.022$     & $\underline{+0.034}$ \\
SCL & $-0.006$ & $-0.008$      & $-0.010$     & $\underline{-0.024}$ \\
\bottomrule
\end{tabular*}
\end{table}

\begin{table}[t]
\caption{Per-subsystem F1 at the F1-maximising threshold (mean over
15 seeds). Best per column in bold.}\label{tab:results-per-subsystem-f1}
\centering
\begin{tabular*}{\tblwidth}{@{\extracolsep{\fill}}lcccc@{}}
\toprule
Variant & RFQ & DTL & CCL & SCL \\
\midrule
Standard    & 0.771 & 0.700 & 0.641 & \textbf{0.744} \\
DS          & 0.774 & 0.713 & 0.633 & 0.737 \\
PW-First    & 0.794 & 0.733 & 0.642 & 0.731 \\
PW-First+SE & \textbf{0.795} & \textbf{0.781} & \textbf{0.658} & 0.719 \\
\bottomrule
\end{tabular*}
\end{table}

\subsection{Per-fault-family results}\label{subsec:analysis-family}

\begin{table*}[t]
\centering
\footnotesize
\caption{Single-channel separability per fault family. Welch's $t$-test compares the per-pulse mean amplitude of normal pulses with that of each fault family, with the pulses of the four subsystems pooled, on the four most discriminative channels. A large $|t|$ and a small $p$ indicate reliably different mean amplitudes. The last two columns give the largest $|t|$ over all 14 channels and the best AUC-ROC obtainable by ranking pulses on the per-pulse mean of one channel ($\mathrm{AUC}_1$), which does not grow with $n$. Rows are ordered by $\mathrm{AUC}_1$, and the horizontal line separates the two strongly separable families from the others. Misc groups all remaining fault names (Section~\ref{subsec:exp-family}).}
\label{tab:welch}
\begin{tabular*}{\tblwidth}{@{\extracolsep{\fill}}lccccccccrcc@{}}
\toprule
  & \multicolumn{2}{c}{B+*IGBT-I} & \multicolumn{2}{c}{C-Flux} & \multicolumn{2}{c}{Mod-V} & \multicolumn{2}{c}{CB-I} & & & \\
Fault family & $t$ & $p$ & $t$ & $p$ & $t$ & $p$ & $t$ & $p$ & $n$ & $\max|t|$ & $\mathrm{AUC}_1$ \\
\midrule
DV/DT   & $-25.86$ & $<10^{-10}$ & $26.18$ & $<10^{-10}$ & $22.84$ & $<10^{-10}$ & $4.08$ & $6.7\times10^{-5}$ & 191 & 26.2 & 0.935 \\
SNS PPS & $-17.31$ & $<10^{-10}$ & $17.03$ & $<10^{-10}$ & $16.73$ & $<10^{-10}$ & $17.35$ & $<10^{-10}$ & 126 & 17.4 & 0.865 \\
\midrule
SCR     & $-6.44$ & $2.1\times10^{-9}$ & $2.82$ & $0.0056$ & $8.40$ & $<10^{-10}$ & $5.03$ & $1.6\times10^{-6}$ & 131 & 8.4 & 0.766 \\
IGBT    & $1.11$ & $0.2708$ & $1.07$ & $0.2871$ & $5.30$ & $1.5\times10^{-6}$ & $3.08$ & $0.0030$ & 66 & 5.3 & 0.764 \\
Driver  & $-6.95$ & $1.4\times10^{-10}$ & $3.16$ & $0.0019$ & $8.22$ & $<10^{-10}$ & $7.07$ & $<10^{-10}$ & 135 & 8.2 & 0.733 \\
Flux    & $-6.18$ & $1.9\times10^{-8}$ & $4.82$ & $5.8\times10^{-6}$ & $6.08$ & $2.8\times10^{-8}$ & $5.86$ & $7.7\times10^{-8}$ & 90 & 6.6 & 0.726 \\
Misc    & $-11.71$ & $<10^{-10}$ & $13.65$ & $<10^{-10}$ & $11.93$ & $<10^{-10}$ & $7.59$ & $<10^{-10}$ & 677 & 14.1 & 0.650 \\
\bottomrule
\end{tabular*}
\end{table*}

\begin{figure}
    \centering
    \includegraphics[width=0.62\linewidth]{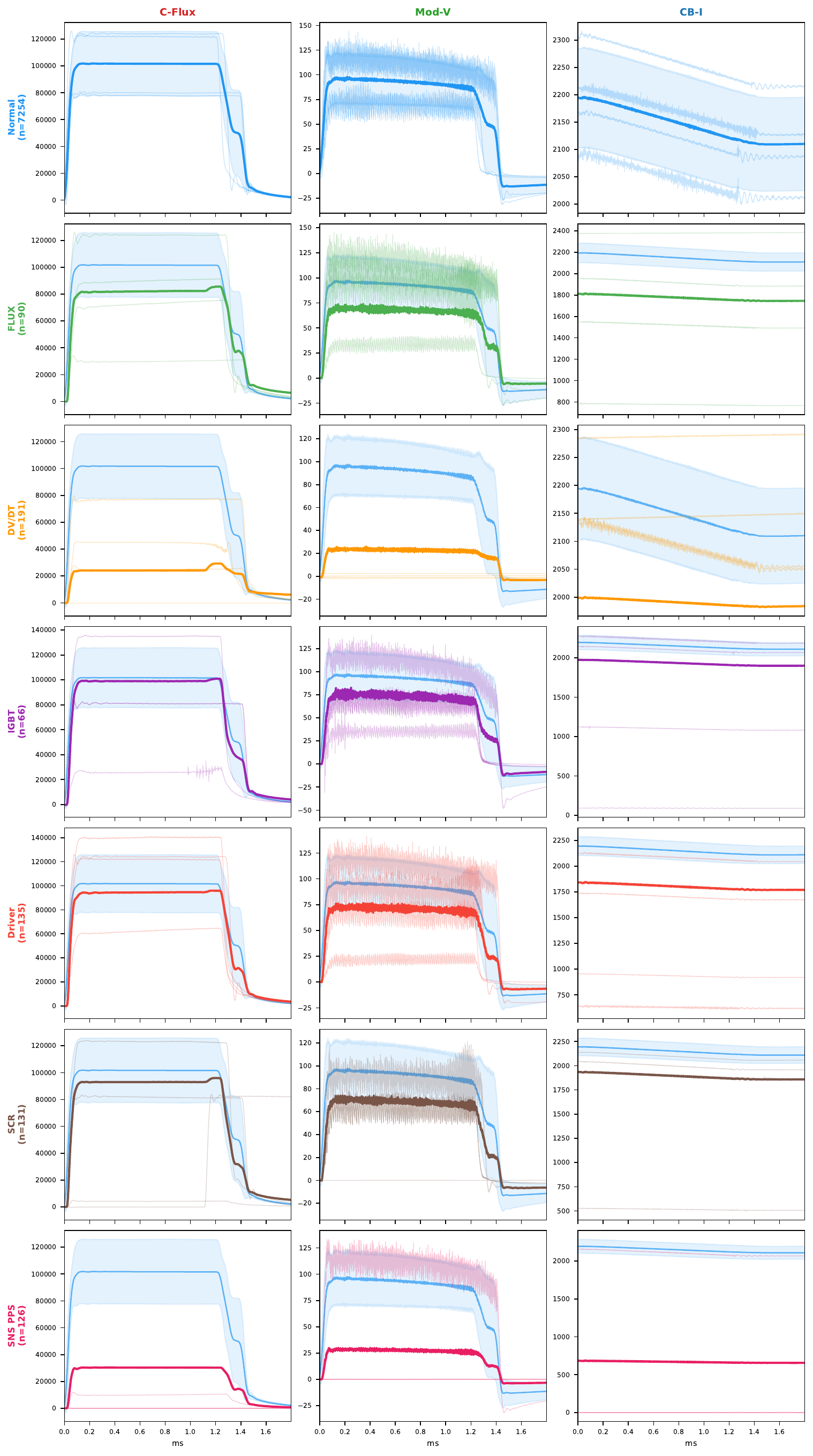}
    \caption{Waveforms of each fault family.}
    \label{fig:type_anomaly}
\end{figure}
\begin{figure}
    \centering
    \includegraphics[width=0.8\textwidth]{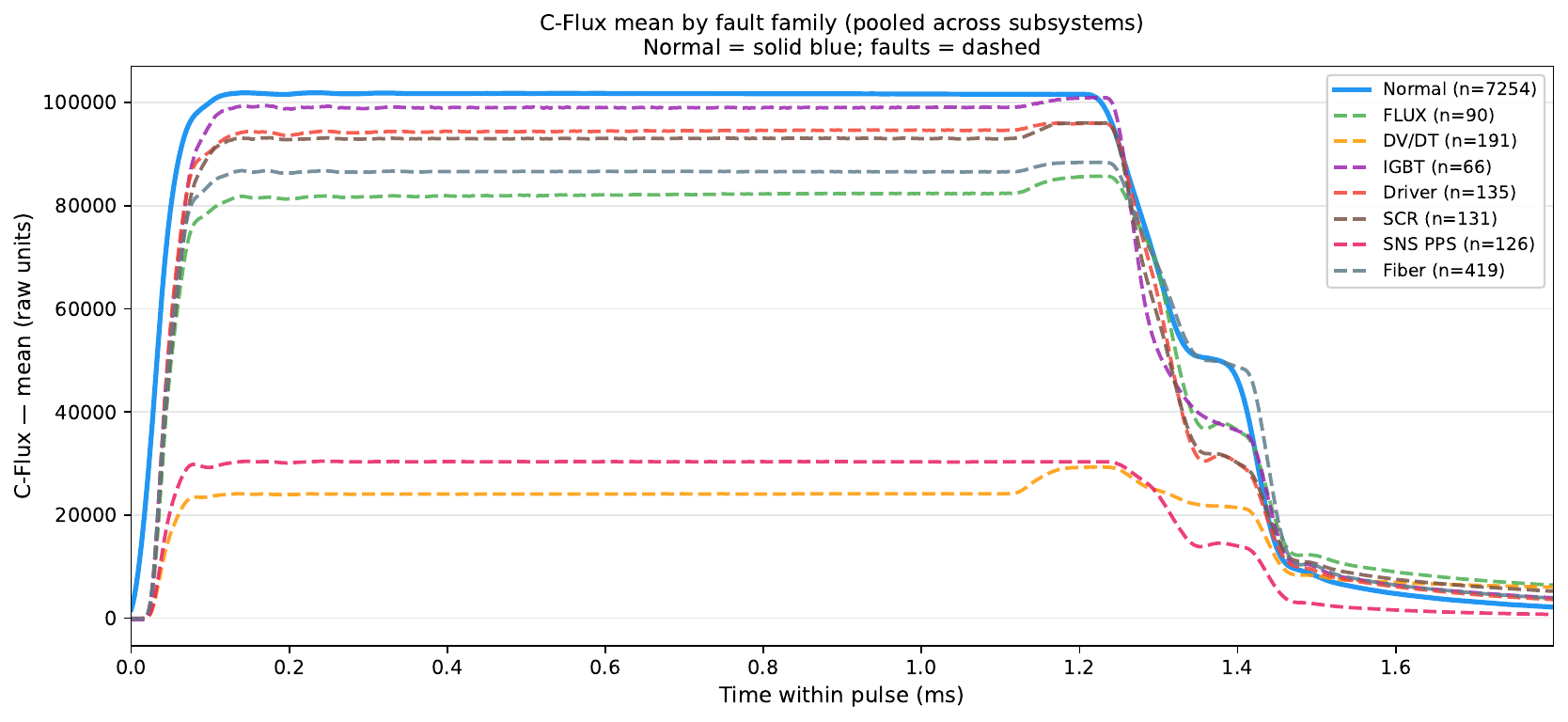}
    \caption{Mean C-Flux trace for each fault family.}
    \label{fig:c-flux}
\end{figure}
\begin{table*}[t]
\centering
\footnotesize
\caption{Per-fault-family detection. Positives are the test pulses of the family and negatives all normal test pulses of the subsystem (Section~\ref{subsec:exp-family}). Left: AUC-ROC, mean over the four subsystems and 15 seeds, next to the single-channel separability $\mathrm{AUC}_1$ of Table~\ref{tab:welch}. Right: the effect of the order, measured as the paired difference PW-First minus DS, averaged over subsystems (descriptive) and within each subsystem (two-sided Wilcoxon over the seeds, uncorrected; bold: $p<0.01$, underlined: $p<0.05$). IGBT is scored on CCL and SCL only, and each of its rows reports that subsystem alone. SNS PPS has no test positives on DTL.}
\label{tab:family}
\begin{tabular*}{\tblwidth}{@{\extracolsep{\fill}}lcccccccccccc@{}}
\toprule
 & & & \multicolumn{4}{c}{AUC-ROC} & \multicolumn{5}{c}{$\Delta$ AUC-ROC, PW-First $-$ DS} \\
\cmidrule(lr){4-7}\cmidrule(lr){8-12}
Family & $n$ & $\mathrm{AUC}_1$ & Standard & DS & PW-First & PW-First+SE & mean & RFQ & DTL & CCL & SCL \\
\midrule
DV/DT   & 191 & 0.935 & 0.974 & 0.974 & 0.979 & 0.981 & $+0.005$ & $\mathbf{+0.033}$ & $0.000$ & $-0.011$ & $-0.002$ \\
SNS PPS & 126 & 0.865 & 0.942 & 0.933 & 0.937 & 0.951 & $+0.004$ & $+0.018$ & -- & $-0.016$ & $-0.003$ \\
\midrule
SCR     & 131 & 0.766 & 0.957 & 0.952 & 0.955 & 0.966 & $+0.003$ & $+0.013$ & $+0.018$ & $-0.016$ & $-0.001$ \\
Driver  & 135 & 0.733 & 0.890 & 0.900 & 0.905 & 0.920 & $+0.005$ & $+0.006$ & $+0.009$ & $+0.004$ & $0.000$ \\
Flux    &  90 & 0.726 & 0.881 & 0.894 & 0.927 & 0.934 & $+0.032$ & $\underline{+0.067}$ & $\mathbf{+0.049}$ & $+0.031$ & $-0.014$ \\
Misc    & 677 & 0.650 & 0.841 & 0.854 & 0.894 & 0.913 & $+0.040$ & $\mathbf{+0.038}$ & $\mathbf{+0.076}$ & $\mathbf{+0.046}$ & $-0.002$ \\
\midrule
IGBT (CCL) & 40 & 0.764 & 0.883 & 0.874 & 0.899 & 0.912 & -- & -- & -- & $\underline{+0.025}$ & -- \\
IGBT (SCL) & 22 & --    & 0.945 & 0.932 & 0.923 & 0.939 & -- & -- & -- & -- & $-0.009$ \\
\bottomrule
\end{tabular*}
\end{table*}

We now ask whether the effect of the order is the same for all fault families or limited to families whose precursors are hard to see on a single channel. Table~\ref{tab:welch} reports the single-channel separability of each family. Two families differ clearly from the rest. DV/DT and SNS PPS shift the main channels far from their normal range, with a largest $|t|$ of 26.2 and 17.4 and an $\mathrm{AUC}_1$ of 0.935 and 0.865. The other families are only weakly separable on any single channel, with $\mathrm{AUC}_1$ between 0.65 and 0.77. Misc has a large $t$ only because it contains 677 pulses, and on $\mathrm{AUC}_1$ it is the least separable group. Figures~\ref{fig:type_anomaly} and~\ref{fig:c-flux} show the waveforms of each family and the mean C-Flux trace per family.

Table~\ref{tab:family} reports the corresponding detection results. For the two strongly separable families, the order of mixing and filtering makes no difference. The average effect is $+0.005$ and $+0.004$ AUC-ROC, all variants, including Standard, exceed 0.93, and within subsystems the effect is not significant in six of the seven cells that can be evaluated. The exception, DV/DT on the RFQ ($+0.033$, $p=0.004$), is based on 3 to 8 positives per test split and does not appear on any other subsystem. When a fault shifts single channels this strongly, filtering each channel on its own at the lowest level, as DS does, is sufficient, and mixing the channels first adds nothing. This agrees with the benchmark studies of Section~\ref{subsec:rw-cicd}, here within a coupled system and for the families whose anomalies are visible on single channels.

The effect of the order appears among the weakly separable families. Flux, the physical family with the lowest $\mathrm{AUC}_1$, gains $+0.032$ AUC-ROC on average when channels are mixed before filtering. The gain is present on the RFQ ($+0.067$, $p=0.011$), the DTL ($+0.049$, $p<0.001$) and the CCL ($+0.031$, $p=0.084$), and absent on the SCL. The Misc group behaves in the same way, with an average gain of $+0.040$ that is significant on the same three subsystems. IGBT faults, for which a cross-channel precursor is most expected on physical grounds, gain $+0.025$ ($p=0.026$) on the CCL, and $+0.038$ ($p=0.009$) with the SE gate, with 6 to 17 positives per test split. Driver and SCR faults are also weakly separable but do not gain from the order ($+0.005$ and $+0.003$ on average, not significant on any subsystem). With the SE gate they gain $+0.020$ and $+0.014$ on average, but the gain is significant in only one subsystem each (Driver on the DTL, $p=0.028$, and SCR on the RFQ, $p=0.023$), so we do not count them as cross-channel gains.

The pattern is therefore asymmetric. Strong single-channel separability rules out a gain from cross-channel modelling. Weak separability is necessary for a gain but does not guarantee it: among the weakly separable families, the gain appears for Flux, Misc and IGBT on the CCL, and not for Driver and SCR. In other words, a CD treatment of the channels helps only where the individual channels do not already reveal the fault. For the strongly separable families this agrees with the benchmark studies, and for the weakly separable ones it shows a gain that those benchmarks could not reveal.

The IGBT results are worth examining in more detail, because the two subsystems where they can be evaluated behave differently. The dataset distinguishes nine IGBT subtypes, each named after the switch whose current it concerns. On the CCL, the three subtypes with at least four pulses show no marginal signal on the channel they are named after ($t=-0.98$, $-1.40$ and $+2.31$ against normal CCL pulses), so they are weakly separable, and there the order helps. On the SCL the order does not help ($-0.009$, $p=0.28$), and the composition of the IGBT pulses explains why. Thirteen of the 22 belong to a single subtype, IGBT B+* Low Fault, whose per-pulse mean on the B+* channel differs from that of normal SCL pulses by $t=-77$, with no overlap between the two ranges. This subtype is strongly separable although the family as a whole is not, and it dominates the SCL estimate. The IGBT results are therefore consistent with the pattern above on both subsystems, and they also show why the two subsystems should not be pooled.

No family gains on the SCL, including Flux and Misc, which gain on every other subsystem. This matches the reversed ranking in Table~\ref{tab:results-per-subsystem-auprc}. The SCL has the largest share of strongly separable pulses (DV/DT and SNS PPS make up 30.8\% of its faults, against 10.5 to 20.3\% on the other subsystems), but this cannot be the whole explanation, since the weakly separable families do not gain there either. The SCL also leaves less room for improvement. Under DS its per-family AUC-ROC averages 0.957 and no family falls below 0.91, against averages of 0.89 to 0.91 on the other three subsystems. Why the order slightly reduces performance on the SCL, instead of leaving it unchanged, remains an open question.

The separability analysis also clarifies the role of Mod-V. Mod-V separates normal and pre-fault pulses significantly in every family of Table~\ref{tab:welch}, including IGBT, which makes it the most broadly informative single channel. This explains why a reconstruction model such as the CVAE, which scores pulses from the Mod-V reconstruction error alone, performs well (Section~\ref{subsec:analysis-detection-hvcm}). The ablations below show, however, that the trained classifiers rely most on C-Flux, not on Mod-V. Section~\ref{subsec:analysis-corr} reconciles the two observations.

\subsection{Channel reliance}\label{subsec:analysis-corr}

The per-family results show where mixing channels first helps. To examine which input channels the trained detectors rely on, we run two ablations on the trained models of all 15 seeds. We distinguish \emph{input channels}, the 14 raw HVCM signals of Section~\ref{subsec:hvcm-data}, from \emph{derived channels}, the $F$ feature maps produced by the pointwise convolution in each block, with $F \in \{32, 64, 128, 64\}$. Both ablations measure reliance on input channels, first for groups of physically related sensors and then for individual channels.

\subsubsection{Sensor-group zero-out}

We first remove groups of physically related channels at test time by setting them to zero. After standardisation, this is equivalent to replacing each channel with its training-set mean. The groups follow the physical layout of the HVCM described in Section~\ref{sec:hvcm}.

\begin{table}[t]
\caption{Mean drop in test AUC-PR when a whole physical sensor group is set to zero at test time. Larger values indicate heavier reliance on that group. Largest drop per variant in bold.}\label{tab:group-drops}
\centering
\begin{tabular*}{\tblwidth}{@{\extracolsep{\fill}}lcccc@{}}
\toprule
Variant & IGBT & DV/DT & Flux & Power \\
\midrule
Standard    & 0.267          & 0.054          & 0.273          & 0.346 \\
DS          & 0.254          & 0.048          & 0.326 & 0.222          \\
PW-First    & 0.252          & 0.036          & 0.233          & 0.321 \\
PW-First+SE & \textbf{0.280}          & \textbf{0.062}          & \textbf{0.346}          & \textbf{0.364} \\
\bottomrule
\end{tabular*}
\end{table}

Table~\ref{tab:group-drops} reports the drops in AUC-PR, averaged over the four subsystems and 15 seeds. Three of the four variants (Standard, PW-First and PW-First+SE) rely most on the Power group, with drops of $0.32$ to $0.36$. DS is the exception. Its largest drop is on the Flux group ($0.326$), and its drop on the Power group is much smaller ($0.222$). It follows that reliance on the Power group is a property of the variants that mix the channels before or together with the temporal filtering. PW-First+SE also shows the largest drop of all variants on every group, including the largest drop in the table ($0.364$, Power). This makes it the variant most sensitive to the removal of any sensor group. Finally, the DV/DT group contributes least for all four variants, with drops between $0.036$ and $0.062$.

\subsubsection{Per-channel zero-out}

We then remove one input channel at a time. For each of the 14 input channels, we set that channel to zero at test time and measure the drop in test AUC-PR relative to the model with all channels.

\begin{table}[t]
\caption{Mean drop in test AUC-PR when a single input channel is set to zero at test time. Larger values indicate heavier reliance on that channel. Largest drop per channel highlighted in bold.}\label{tab:channel-drops}
\centering
\renewcommand{\arraystretch}{1.05}
\begin{tabular*}{\tblwidth}{@{\extracolsep{\fill}}lcccc@{}}
\toprule
Channel & Standard & DS & PW-First & PW-First+SE \\
\midrule
A+      & \textbf{0.076} & 0.053          & 0.040          & 0.059          \\
A+*     & 0.039          & 0.027          & 0.037          & \textbf{0.047} \\
B+      & 0.046          & 0.035          & \textbf{0.052} & 0.048          \\
B+*     & \textbf{0.083} & \textbf{0.083} & 0.052          & 0.057          \\
C+      & 0.038          & \textbf{0.043} & 0.039          & 0.039          \\
C+*     & \textbf{0.070} & 0.054          & 0.056          & 0.067          \\
DV/DT   & 0.062          & 0.057          & 0.031          & \textbf{0.064} \\
A-Flux  & \textbf{0.063} & 0.049          & 0.032          & 0.048          \\
B-Flux  & \textbf{0.062} & 0.054          & 0.041          & \textbf{0.062} \\
C-Flux  & \textbf{0.275} & 0.217          & 0.240          & 0.246          \\
Mod-V   & \textbf{0.209} & 0.164          & 0.177          & 0.181          \\
Mod-I   & 0.005          & 0.019          & 0.028          & \textbf{0.051} \\
CB-I    & 0.241          & 0.211          & \textbf{0.283} & 0.168          \\
CB-V    & 0.037          & 0.038          & 0.039          & \textbf{0.050} \\
\bottomrule
\end{tabular*}
\end{table}

Table~\ref{tab:channel-drops} reports the mean drop per channel over the four subsystems and 15 seeds (60 evaluations per cell). Three observations follow. First, no channel has a negative mean drop for any variant. The smallest contributions for PW-First come from Mod-I and DV/DT. DV/DT is the time derivative of the modulator voltage, so the model may recover much of its information from Mod-V when it is removed. Second, three channels, C-Flux, Mod-V and CB-I, dominate, with drops between $0.16$ and $0.28$, roughly four to six times larger than those of the other channels. C-Flux is the most informative channel for three of the four variants, in line with \citet{radaideh2022dsp}, who trained their autoencoders on C-Flux alone in the RFQ and reported an AUC-ROC of up to $0.90$. Third, DS relies less on each of the top channels than the other variants, with a C-Flux drop of $0.217$ against $0.240$ to $0.275$. This is the per-channel counterpart of the group-level finding that DS distributes its reliance differently across sensor groups.

The per-channel results also explain the apparent contradiction between the separability and ablation analyses. Mod-V is the most broadly separating channel in Table~\ref{tab:welch}, but C-Flux is the channel the classifiers rely on most in Table~\ref{tab:channel-drops}. The two analyses measure different things. The Welch statistic measures whether a channel separates the classes on its own, whereas the ablation measures whether the model can compensate for a missing channel with the others. The correlation analysis of Figure~\ref{fig:correlation_study} connects the two. In both the RFQ and the CCL, the channel pair whose correlation changes most between normal and pre-fault pulses is C-Flux and Mod-V. The two channels are coupled, and their coupling changes before a fault. One possible explanation is that part of the information in Mod-V is also available through C-Flux, so the models can rely on C-Flux while Mod-V remains the strongest single-channel separator. This suggests that some of the most fault-sensitive information lies in the relationship between the two channels, which is the kind of dependency that mixing channels before temporal filtering can capture.

\subsection{Comparison with published HVCM anomaly detectors}
\label{subsec:analysis-detection-hvcm}

The comparisons above support the conclusions of the paper. This section places the detectors alongside the published HVCM detectors of Section~\ref{subsec:exp-baselines}, whose results are taken from the original papers.

\begin{table*}[t]
\caption{Comparison with prior work on the RFQ subsystem of the public HVCM dataset \citep{radaideh2022data}. Mean values from the original papers, where available. ``--'' indicates that the metric is not reported in the source paper. AE: autoencoder. Best per column in bold.}\label{tab:comparison-rfq}
\centering
\begin{tabular*}{\tblwidth}{@{\extracolsep{\fill}}llccccccc@{}}
\toprule
Method & Source & Acc & AUC-ROC & Prec & Rec & F1 & G-Mean \\
\midrule
\multicolumn{8}{l}{\emph{Unsupervised baselines} (RFQ, C-Flux only)} \\
\quad LSTM-AE     & \citet{radaideh2022dsp} & 0.870 & 0.900 & --$^{\dagger}$ & --$^{\dagger}$ & --$^{\dagger}$ & -- \\
\quad GRU-AE      & \citet{radaideh2022dsp} & 0.850 & 0.890 & --$^{\dagger}$ & --$^{\dagger}$ & --$^{\dagger}$ & -- \\
\quad ConvLSTM-AE & \citet{radaideh2022dsp} & 0.850 & 0.890 & --$^{\dagger}$ & --$^{\dagger}$ & --$^{\dagger}$ & -- \\
\midrule
\multicolumn{8}{l}{\emph{Supervised baselines} (14 channels)} \\
\quad KNN     & \citet{zhou2024multisensor} & 0.871 & 0.90 & 0.755 & 0.574 & 0.649 & 0.660 \\
\quad RF      & \citet{zhou2024multisensor} & 0.868 & 0.92 & 0.704 & 0.619 & 0.647 & 0.658 \\
\quad SVM     & \citet{zhou2024multisensor} & 0.871 & 0.91 & \textbf{0.930} & 0.425 & 0.581 & 0.629 \\
\quad CNN     & \citet{zhou2024multisensor} & 0.841 & 0.88 & 0.919 & 0.387 & 0.539 & 0.596 \\
\quad LSTM    & \citet{zhou2024multisensor} & 0.863 & 0.91 & 0.825 & 0.549 & 0.644 & 0.673 \\
\quad CNN+LSTM & \citet{zhou2024multisensor} & 0.904 & 0.93 & 0.844 & 0.655 & 0.721 & 0.744 \\
\midrule
\multicolumn{8}{l}{\emph{This work} (14 channels)} \\
\quad Standard    & ours & 0.915 & 0.882 & 0.878 & 0.692 & 0.771 & 0.819 \\
\quad DS          & ours & 0.914 & 0.904 & 0.874 & 0.705 & 0.774 & 0.825 \\
\quad PW-First    & ours & \textbf{0.917} & 0.933 & 0.835 & 0.764 & 0.794 & 0.854 \\
\quad PW-First+SE & ours & \textbf{0.917} & \textbf{0.938} & 0.825 & \textbf{0.773} & \textbf{0.795} & \textbf{0.858} \\
\bottomrule
\end{tabular*}
\smallskip\\
{\footnotesize $^{\dagger}$ \citet{radaideh2022dsp} use an inverted convention in which the positive class is ``Run'' (normal), so their Precision and Recall correspond to specificity-style quantities on the Fault class and are not directly comparable to the values reported here. We omit these cells to avoid a misleading direct comparison. Their AUC and Accuracy use the conventional definitions and are reported as published.}
\end{table*}

\begin{table*}[t]
\caption{Per-subsystem comparison against the supervised CNN+LSTM model of
\citet{zhou2024multisensor} on the four HVCM subsystems. Only seed-averaged values tabulated in the source are used. Best per column in
bold.}\label{tab:comparison-zhou}
\centering
\begin{tabular*}{\tblwidth}{@{\extracolsep{\fill}}lcccccccc@{}}
\toprule
 & \multicolumn{2}{c}{RFQ} & \multicolumn{2}{c}{DTL} & \multicolumn{2}{c}{CCL} & \multicolumn{2}{c}{SCL} \\
\cmidrule(lr){2-3}\cmidrule(lr){4-5}\cmidrule(lr){6-7}\cmidrule(lr){8-9}
Method & F1 & Acc & F1 & Acc & F1 & Acc & F1 & Acc \\
\midrule
CNN+LSTM $^{\dagger}$   & 0.721 & 0.904 & --    & 0.927 & --    & \textbf{0.925} & --    & 0.922 \\
\midrule
Standard    & 0.771 & 0.915 & 0.700 & 0.928 & 0.641 & 0.908 & \textbf{0.744} & \textbf{0.926} \\
DS          & 0.774 & 0.914 & 0.713 & 0.928 & 0.633 & 0.906 & 0.737 & 0.924 \\
PW-First    & 0.794 & \textbf{0.917} & 0.733 & 0.930 & 0.642 & 0.893 & 0.731 & 0.920 \\
PW-First+SE & \textbf{0.795} & \textbf{0.917} & \textbf{0.781} & \textbf{0.941} & \textbf{0.658} & 0.903 & 0.719 & 0.918 \\
\bottomrule
\end{tabular*}
\smallskip\\
{\footnotesize $^{\dagger}$ Seed averages tabulated in Table~5 (RFQ) and
Table~6 (Acc only, all four subsystems) of \citet{zhou2024multisensor}. F1 for
DTL, CCL, SCL is not tabulated.}
\end{table*}

\begin{table*}[t]
\caption{Per-fault-family AUC-ROC comparison with the multi-module CVAE of \citet{alanazi2023cvae}, on the five families with a stable per-subsystem estimate. The CVAE values are read from Figure 8 of the original paper. Our values are the mean of the four per-subsystem models. Best per column in bold.}\label{tab:comparison-cvae-perfault}
\centering
\begin{tabular*}{\tblwidth}{@{\extracolsep{\fill}}llccccc@{}}
\toprule
Method & Source & DV/DT & Flux & Driver & SCR & SNS PPS \\
\midrule
Multi-module CVAE$^{\dagger}$ & \citep{alanazi2023cvae} & \textbf{0.98} & 0.85 & 0.83 & 0.93 & \textbf{0.96} \\
\midrule
Standard    & ours & 0.974          & 0.881                  & 0.890          & 0.957          & 0.942 \\
DS          & ours & 0.974          & 0.894                  & 0.900          & 0.952          & 0.933 \\
PW-First    & ours & 0.979          & 0.927                  & 0.905          & 0.955          & 0.937 \\
PW-First+SE & ours & \textbf{0.981} & \textbf{0.934} & \textbf{0.920} & \textbf{0.966} & 0.951 \\
\bottomrule
\end{tabular*}
\smallskip\\
{\footnotesize $^{\dagger}$ \citet{alanazi2023cvae} use the Mod-V reconstruction error alone as the anomaly score for every fault family, so this row reflects single-channel performance under post-hoc channel selection, reported as single-run point estimates without seed averaging.}
\end{table*}

Three studies report anomaly detection results on the public HVCM dataset, with different protocols. \citet{radaideh2022dsp} use a single channel (C-Flux) of the RFQ subsystem and unsupervised reconstruction. \citet{alanazi2023cvae} use all 14 channels and all 15 modulators, also without supervision, with a CVAE. \citet{zhou2024multisensor} use the same supervised setting as this paper, with 14 channels and four subsystems.

Table~\ref{tab:comparison-rfq} collects the published results on the RFQ, the only subsystem reported in all three studies. PW-First+SE outperforms the strongest published RFQ detector, the supervised CNN+LSTM of \citet{zhou2024multisensor}, on F1, recall, G-Mean and AUC-ROC, with very similar accuracy. The unsupervised autoencoders of \citet{radaideh2022dsp} reach an AUC-ROC of up to 0.90 on the RFQ, about 0.04 below PW-First+SE.

For the other three subsystems (Table~\ref{tab:comparison-zhou}), the comparison is limited by what \citet{zhou2024multisensor} report: F1 only for the RFQ, and accuracy for all four subsystems. On the RFQ, PW-First+SE improves F1 over the CNN+LSTM ($0.795$ against $0.721$). In accuracy, PW-First+SE is higher on the RFQ ($0.917$ against $0.904$) and the DTL ($0.941$ against $0.927$), and lower on the CCL ($0.903$ against $0.925$) and the SCL ($0.918$ against $0.922$). These differences are small, and since accuracy is dominated by the majority class, it does not discriminate well between detectors on this dataset. We therefore take Table~\ref{tab:comparison-zhou} to show that our detectors are competitive with the strongest published supervised detector on F1, the metric that reflects fault detection, not that they are uniformly more accurate.

Three caveats apply. First, the precision and recall of \citet{radaideh2022dsp} are omitted, because they treat the Run class as positive, which reverses the meaning of these metrics for the Fault class. Second, \citet{alanazi2023cvae} report AUC-ROC by fault family and not by subsystem, and their values are read from figures, so they are approximate. Third, our threshold-dependent metrics are computed at each model's F1-maximising threshold, so they are an upper bound on what a fixed operating threshold would achieve, whereas the published values use the thresholds chosen in the original papers.

Table~\ref{tab:comparison-cvae-perfault} compares the per-family AUC-ROC with that of the CVAE of \citet{alanazi2023cvae} for the five families with a stable per-subsystem estimate. IGBT is omitted for the coverage reason given in Section~\ref{subsec:exp-family}, and its results are in Table~\ref{tab:family}. PW-First+SE matches or exceeds the CVAE on the families the CVAE finds hardest. The CVAE scores every family from the Mod-V reconstruction error alone, and its ranking of the families, best on DV/DT and SNS PPS and worst on Flux and Driver, broadly follows the single-channel separability of Table~\ref{tab:welch}, as expected for a score based on a single channel.

\section{Conclusion}\label{sec:conclusion}

This paper examined how one architectural decision in a CNN, the order in which channels are mixed and time is filtered, affects the way a multivariate time series anomaly detector captures the dependencies between channels, and hence its detection performance, on a sensor system whose channels are physically coupled. Public benchmarks cannot answer this question because their anomalies are mostly visible on single channels. The HVCMs of the SNS, with fourteen coupled sensors, fault labels by type and public data, make it possible to study it. We compared convolutional detectors that differ only in this order, from low-level processing close to a CI model to processing that is CD from the first operation.

The benefit of cross-channel modelling depends on the fault type. Families whose precursors shift a single channel far from its normal range, DV/DT and SNS PPS, are detected equally well whatever the order, which reproduces within a coupled system the result reported on public benchmarks. For families with weak single-channel separability the order matters. Mixing channels before filtering increases the per-family AUC-ROC by 0.03 to 0.04 for Flux and for the Misc group on three of the four subsystems, and by 0.025 to 0.038 for IGBT faults on the CCL, where the IGBT subtypes show no marginal signal on the channels they are named after. Two weakly separable families, Driver and SCR, do not benefit, so weak single-channel separability is a necessary but not a sufficient condition for a gain. 

At the subsystem level, the largest improvement on the RFQ, DTL and CCL also comes from the change of order, the depthwise factorisation alone has no effect, and the SE gate adds a further improvement on the DTL. The resulting detector, PW-First+SE, reaches an average AUC-PR of 0.816 and AUC-ROC of 0.934 with 1.9 times fewer parameters than the Standard baseline, and is competitive with published HVCM detectors on the metrics that can be compared. On the SCL, no fault family benefits from the order, including the weakly separable ones. Its fault composition and its high baseline performance explain part of this result, but not why the order slightly reduces performance there.

For the operation of the HVCMs and of similar monitored systems, the results have practical consequences. Faults that are clearly visible on a single sensor, such as the DV/DT and SNS PPS families, are detected equally well by all variants, and even the mean amplitude of a single channel separates DV/DT faults from normal pulses with an AUC-ROC of 0.935, so a multi-channel model adds little for them. Faults that are weakly visible on any single sensor, such as Flux faults, are where a detector that combines channels before filtering them in time detects more, and where a multi-channel model is therefore worth its cost. The channel ablations also show which sensors fault detection depends on. Removing C-Flux, Mod-V or CB-I reduces AUC-PR by 0.16 to 0.28, four to six times more than removing any other channel, so the performance of the detector depends on these three sensors remaining available and well calibrated. Finally, the way the detector treats the channels can be changed by a local modification of the convolutional block, without redesigning the rest of the model. Since no single choice was best on every subsystem, this choice is best made separately for each subsystem, on validation data.

Beyond this application, the results have two implications for the choice between CI and CD detectors. First, the benchmark finding that CD models offer no gain is not a general property of multivariate anomalies. A single coupled system already contains fault families for which the gain is present, so the finding is better read as a statement about the anomalies that the benchmarks contain. Second, whether a CD treatment of the channels helps depends on how visible each fault is on individual channels, so evaluations of CI and CD detectors should report results by fault type and not only in aggregate. Testing whether the same pattern holds in other coupled systems will require further datasets with fault-type labels.

\section*{Acknowledgements}

The authors thank the Spallation Neutron Source team at Oak Ridge National Laboratory for releasing the public HVCM dataset used in this study \cite{radaideh2022data}. The SNS is a DOE Office of Science User Facility operated by Oak Ridge National Laboratory. The authors would also like to thank M. Weber, Director of the IFMIF-DONES particle accelerator being built in Granada, Spain, for feedback and insights on anomalies in systems of this kind.

\section*{Statements and Declarations}

\begin{itemize}

\item \textbf{Funding:} This work was supported by the Spanish Ministry for Digital Transformation and the Civil Service through the Recovery, Transformation and Resilience Plan, funded by the European Union NextGenerationEU [grant number TSI-100927-2023-1].

\item \textbf{Competing interests:} The authors have no competing interests to declare that are relevant to the content of this article.

\item \textbf{Data availability:} The HVCM dataset analysed in this study is openly available on Mendeley Data (\url{https://data.mendeley.com/datasets/kbbrw99vh8/5}) under a CC~BY~4.0 licence \cite{radaideh2022data}. The code that produces every result in this paper is available at \url{https://github.com/adiazcencillo/hvcm-fault-detection}.

\item \textbf{Author contribution:} Alberto D. Cencillo: Conceptualisation, Methodology, Software, Investigation, Formal analysis, Writing – review \& editing.

Leonardo Concepción: Conceptualisation, Methodology, Writing – original draft, Writing – review \& editing.

Julián Luengo: Methodology, Supervision, Writing – review \& editing, Funding acquisition.

Isaac Triguero: Methodology, Supervision, Writing – review \& editing, Funding acquisition.

\item \textbf{Declaration of generative AI use:} During the preparation of this work the authors used Claude (Anthropic) in order to assist with drafting, editing and restructuring the text. After using this tool, the authors reviewed and edited the content as needed and take full responsibility for the content of the published article.

\end{itemize}

\bibliographystyle{unsrtnat}
\bibliography{sn-bibliography}

\end{document}